\documentclass[10pt,twocolumn,letterpaper]{article}

\usepackage{cvpr}              % To produce the CAMERA-READY version
\usepackage[dvipsnames]{xcolor}

\definecolor{cvprblue}{rgb}{0.21,0.49,0.74}
\usepackage[pagebackref,breaklinks,colorlinks,citecolor=cvprblue,bookmarks=false]{hyperref}
\usepackage[accsupp]{axessibility}
\usepackage{multirow}
\usepackage{algorithm}
\usepackage{algorithmic}
\def\paperID{9625} % *** Enter the Paper ID here
\def\confName{CVPR}
\def\confYear{2024}

\title{CURSOR: Scalable Mixed-Order Hypergraph Matching with CUR Decomposition}

\author{Qixuan Zheng$^{1}$\quad Ming Zhang$^{2}$\thanks{Corresponding author.}\quad Hong Yan$^{1}$\\
$^1$City University of Hong Kong \\
$^2$Hong Kong Applied Science and Technology Research Institute (ASTRI) \\
{\tt\small \{qixuzheng2-c, mzhang367-c\}@my.cityu.edu.hk, h.yan@cityu.edu.hk} 
}

\begin{document}
\maketitle
\begin{abstract}
To achieve greater accuracy, hypergraph matching algorithms require exponential increases in computational resources. Recent kd-tree-based approximate nearest neighbor (ANN) methods, despite the sparsity of their compatibility tensor, still require exhaustive calculations for large-scale graph matching. This work utilizes CUR tensor decomposition and introduces a novel cascaded second and third-order hypergraph matching framework (CURSOR) for efficient hypergraph matching. A CUR-based second-order graph matching algorithm is used to provide a rough match, and then the core of CURSOR, a fiber-CUR-based tensor generation method, directly calculates entries of the compatibility tensor by leveraging the initial second-order match result. This significantly decreases the time complexity and tensor density. A probability relaxation labeling (PRL)-based matching algorithm, specifically suitable for sparse tensors, is developed. Experiment results on large-scale synthetic datasets and widely-adopted benchmark sets demonstrate the superiority of CURSOR over existing methods. The tensor generation method in CURSOR can be integrated seamlessly into existing hypergraph matching methods to improve their performance and lower their computational costs.
\end{abstract}    
\section{Introduction}
\label{sec:intro}

\begin{figure}[htbp]
\centering
\subfloat[Traditional ANN-based hypergraph matching]{\includegraphics[width=0.49\textwidth]{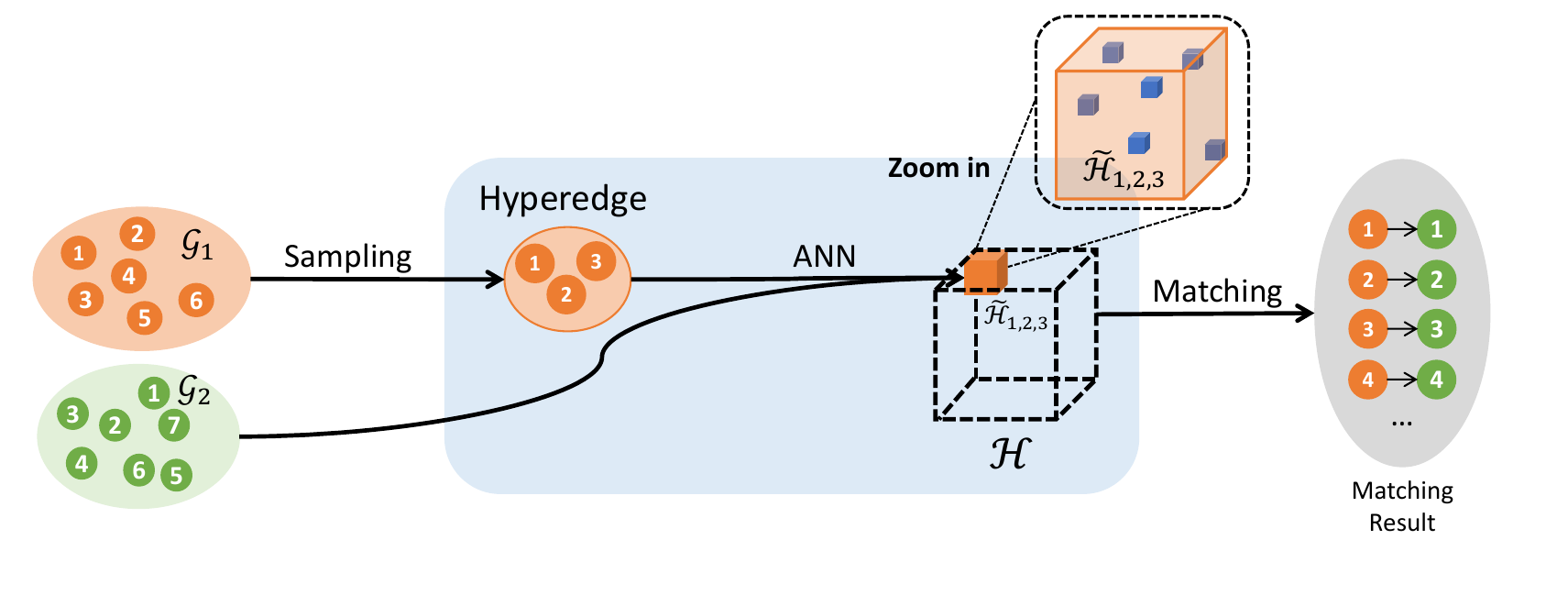}\label{subfig:traditional}}\\
 \subfloat[The proposed CURSOR with PRL-based matching]{\includegraphics[width=0.49\textwidth]{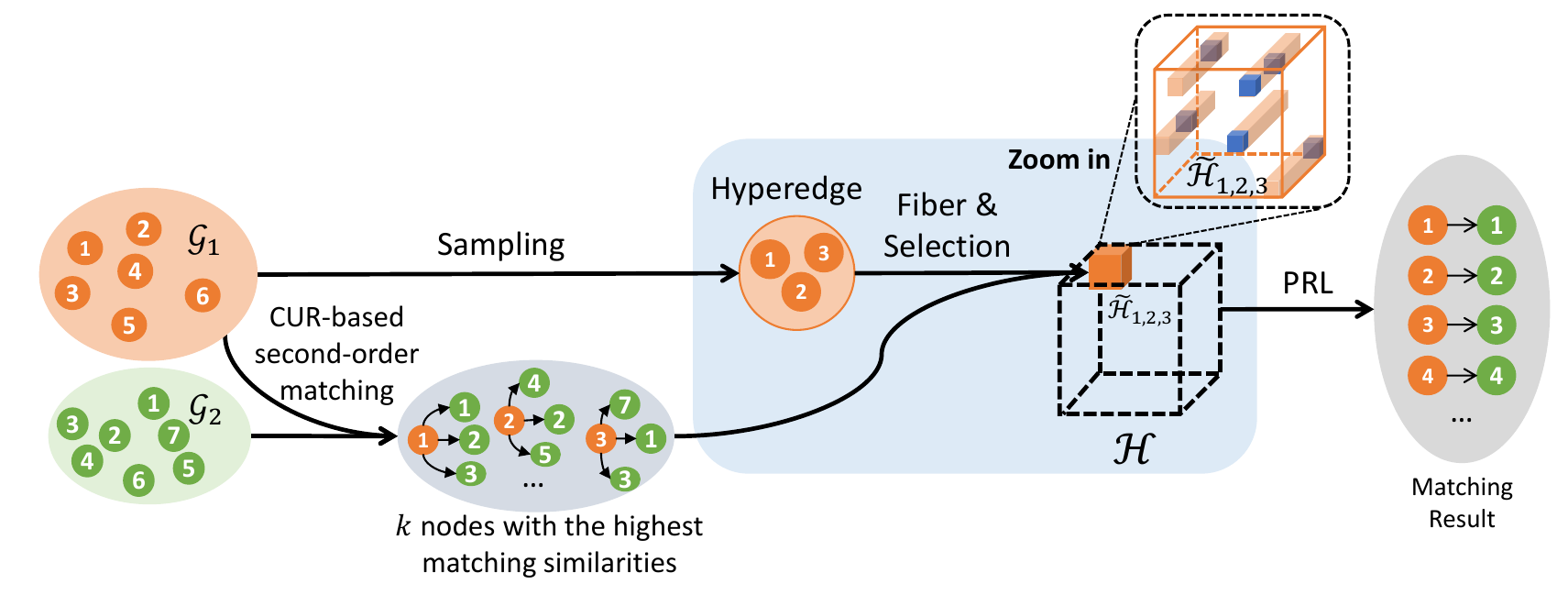}\label{subfig:curann}}
\caption{The comparison between the traditional ANN-based framework and CURSOR. Instead of calculating the whole tensor block (the light orange area in \subref{subfig:traditional}) and extracting the highest compatibilities in each block (the blue cubes), CURSOR only calculates a small number of block fibers (the light orange area in \subref{subfig:curann}) and retains fewer elements in these fibers, effectively reducing computational costs for large-scale hypergraph matching. The method chooses the fibers based on the second-order matching result. CURSOR calculates fibers in all three tensor modes, and only one is shown in \subref{subfig:curann} for clarity.}
\label{fig:method}
\end{figure}

Finding the correspondences between a pair of feature sets by graph-matching has many applications in computer vision and pattern recognition tasks like feature tracking \cite{he2021learnable, zhou2018online, li2023distributed}, image classification \cite{wu2021gm}, object detection \cite{li2022sigma}, and gene-drug association identification \cite{chen2019hogmmnc}. The second-order graph matching (pairwise matching) problem is a quadratic assignment problem (QAP), which is NP-hard \cite{lawler1963quadratic}. Efforts to find soft-constraint approximate solutions \cite{leordeanu2005spectral, cour2006balanced, cho2010reweighted, wang2019functional} have been limited to the representation of pairwise compatibilities.\par 
%\begin{figure}
%    \centering
%    \includegraphics[width=0.5\textwidth]{figure/flow_chart.pdf}
%    \caption{The flow chart of the proposed cascaded hypergraph matching framework. It first calculates a rough matching result with the CUR-based graph matching algorithm. Then, the tensor block fibers corresponding to the second-order matching result will be calculated to form the sparse compatibility tensor. The PRL-based matching algorithm is applied to the tensor for the final matching result.}
%    \label{flowchart}
%\end{figure}

Higher-order graphs, known as hypergraphs \cite{duchenne2011tensor,lee2011hyper,le2017alternating, nguyen2015flexible, hou2023game}, integrate better geometric information and handle transformations like scaling and rotation better. The hypergraph matching problem considers the compatibility between two hypergraphs as a high-order tensor. The objective function aims to find the maximization over all permutations of the features. A kd-tree-based approximate nearest neighbor (ANN) method \cite{duchenne2011tensor} to reduce the space and time complexity of the hypergraph matching algorithm has been used in many hypergraph matching algorithms (e.g. RRWHM \cite{lee2011hyper}, BCAGM \cite{nguyen2015flexible} and ADGM \cite{le2017alternating}). 
ANN-based methods compute the sparse compatibility tensor by searching for nearest neighbors between randomly sampled hyperedges in the source graph and all the hyperedges in the target graph. Large-scale $k^{\text{th}}$-order hypergraph matching with $n_2$ points in the target graph has $O(n_2^k)$ time and space complexity for the compatibilities of each source graph hyperedge. Achieving a higher matching accuracy requires as many of the highest compatibilities as possible to increase the probability of finding the ground truth paired hyperedges, resulting in a denser compatibility tensor. 
%Therefore, current hypergraph matching algorithms are infeasible for large-scale hypergraph matching tasks, like the interaction predictions between large proteins with thousands of atoms on the molecular surface.

To address the above scalability issue, this work proposes a novel scalable hypergraph matching framework, CURSOR, based on cascaded mixed-order models with CUR decomposition. %CURSOR constructs a sparser compatibility tensor with our proposed fiber-CUR-based tensor generation method and the rough pairwise matching result. 
The comparison of CURSOR and traditional ANN-based methods is illustrated in Fig.~\ref{fig:method}. The traditional ANN-based methods construct the compatibility tensor directly with the nearest neighbors between the randomly sampled source hyperedges and all target ones. CURSOR first computes a roughly intermediate result with the proposed CUR-based second-order matching algorithm, drastically reducing the memory footprint of the compatibility matrices for large-scale graph matching. Subsequently, a small-scale target hyperedge subset, represented as third-order fiber tensors, can be generated utilizing the second-order results to calculate the compatibility tensor, substantially decreasing the computation complexity. %We name the framework CURSOR after the core fiber-CUR-based tensor generation method. 
%Experiment results show that our proposed method can effectively find essential non-zero compatibilities in the sparse compatibility tensor with fewer computation costs and superior matching accuracy. 
The tensor generation method in CURSOR can integrate seamlessly into almost all existing state-of-the-art hypergraph matching algorithms \cite{duchenne2011tensor, lee2011hyper, nguyen2015flexible, le2017alternating, hou2023game} and significantly increase their matching performance at lower computational cost.\par %Furthermore, we develop a probability relaxation labeling (PRL)-based hypergraph matching algorithm to exploit the high tensor sparsity. Experimental results show that the proposed matching algorithm reaches superior matching accuracy.\par
%\subsection{Subsection Heading Here}
The contributions of this work are:
\begin{itemize}
    \item We propose a novel cascaded second and third-order CUR-based hypergraph matching framework, CURSOR, to deal with large-scale problems. Under the same memory limitations, CURSOR can handle a more than ten times larger-scale hypergraph matching problem than current state-of-the-art algorithms.
    \item CURSOR contains a fiber-CUR-based compatibility tensor generation method using the rough matching result from the CUR-based second-order graph matching algorithm, which efficiently decreases the computational complexity and selects the proper sparse tensor entries.
    %\item We propose a cascaded second and third-order hypergraph matching framework to deal with large-scale hypergraph matching problems, which can deal with thousands of points, more than ten times larger than the state-of-the-art algorithms with \red{the same computing resources}.
    %\item We propose a new fiber-CUR-based compatibility tensor generation method with the rough matching result from the CUR-based second-order graph matching algorithm, which efficiently decreases the computation complexity and selects the proper sparse tensor entries.
    \item A PRL-based tensor matching algorithm is developed to significantly accelerate convergence during the matching process and increase the accuracy of matching.
    \item Experiment results show that CURSOR provides state-of-the-art matching accuracy by effectively finding the essential non-zero entries in the compatibility tensor. 
\end{itemize}

\section{Related Works}
\subsection{Hypergraph Matching}

Second-order graph-matching algorithms \cite{leordeanu2005spectral, cour2006balanced, cho2010reweighted, wang2019functional, dupe2023kernelized} represented the geometric consistency between a pair of features as the edges of a graph to avoid ambiguities like repeated patterns and textures. These algorithms pursued an approximate solution as the problem is known to be NP-hard. Among them, Leordeanu and Herbert \cite{leordeanu2005spectral} and Cour \textit{et al.} \cite{cour2006balanced} estimated the rank-1 approximation of the compatibility matrix with power iteration as the flattening of the assignment matrix. Cho \textit{et al.} \cite{cho2010reweighted} provided a novel random walk view for graph matching with a reweighting jump scheme and reduced the constraints in its iterative process. Recently, Wang \textit{et al.} \cite{wang2019functional} proposed a functional representation for graph matching with the additional goal of avoiding the compatibility matrix construction. The performance of second-order methods was limited with the restriction to the normal graph embedding pairwise relationships.\par

In the past decade, to overcome the limitation of pairwise similarity, various hypergraph matching algorithms \cite{duchenne2011tensor, lee2011hyper, le2017alternating, hou2023game} were developed based on the structural compatibilities between the higher-order hyperedges of two hypergraphs. In order to avoid the enormous computational cost of a full compatibility tensor, Duchenne \textit{et al.} \cite{duchenne2011tensor}, extending the SM algorithm proposed by \cite{leordeanu2005spectral} to a higher order, constructed the sparse compatibility tensor with an ANN-based method, which is frequently used for generating compatibility tensors in later hypergraph matching works. Lee \textit{et al.} \cite{lee2011hyper} introduced the method of \cite{cho2010reweighted} to reweighted walk hypergraph matching problems, enforcing the matching constraint with a bi-stochastic normalization scheme. L\^{e}-Huu and Paragios \cite{le2017alternating} decomposed the problem under different constraints and handled it with an alternating direction method of multipliers. Khan \textit{et al.} \cite{khan2019image} directly applied CUR decomposition to the full compatibility matrix and tensor at the cost of a higher space complexity than the ANN-based method. %In the past decade, few works have worked on optimizing the compatibility tensor generation method of the hypergraph matching problems. 
Although the ANN-based method can significantly decrease the time complexity during the matching process, it still has an enormous computational cost to calculate the compatibility tensor. \par 

With the rise of deep learning, various learning-based graph-matching algorithms were proposed to learn the parameters as deep feature extraction hierarchies in a data-driven way \cite{zanfir2018deep, wang2020learning, liu2023graph, cai2023htmatch, sarlin2020superglue, jiang2022graph, Chen_2021_ICCV}. Inspired by Wang \textit{et al.} \cite{wang2019learning}, Liao \textit{et al.} \cite{liao2021hypergraph} proposed the first unified hypergraph neural network, HNN-HM. Unlike the sparse compatibility tensor widely used in ANN-based algorithms, the dense deep feature matrices or tensors require more computation resources. Therefore, compared with ANN-based algorithms, the learning-based hypergraph matching algorithms can only handle much smaller-scale problems under the same memory constraint.\par 

\subsection{CUR Matrix and Tensor Decomposition} 
CUR decomposition \cite{mahoney2009cur} is used to compute the low-rank approximation of a matrix with the actual rows and columns. Assume a matrix $\mathbf{A}\in\mathbb{R}^{m\times n}$. By selecting $c$ columns and $r$ rows from $\mathbf{A}$ as $\mathbf{C}\in\mathbb{R}^{m\times c}$ and $\mathbf{R}\in\mathbb{R}^{r\times n}$, the low-rank approximation of $\mathbf{A}$ can be formulated as $\mathbf{C}(\mathbf{C}^\dagger\mathbf{A}\mathbf{R}^\dagger)\mathbf{R}$, where $\dagger$ is the pseudoinverse and $\mathbf{C}^\dagger\mathbf{A}\mathbf{R}^\dagger$ represents the matrix $\mathbf{U}$. Cai \textit{et al.} \cite{cai2021robust} showed that if $\text{rank}(\mathbf{A})< \min\{c, r\}$, $\mathbf{U}$ can be directly represented by the intersection of $\mathbf{C}$ and $\mathbf{R}$ as its pseudoinverse. Xu \textit{et al.} \cite{xu2015cur} proposed CUR+ to calculate the matrix $\mathbf{U}$ with randomized matrix entries instead of the whole matrix when the matrix is not low rank. For tensor CUR decomposition, randomized fiber CUR decomposition can extend the CUR decomposition to the tensor \cite{cai2021mode}. The method first samples $c_l$ fibers on mode-$l$ and expands the fibers as matrix $\mathbf{C}_l$ along mode-$l$. The intersection of all samples forms tensor $\mathcal{R}$ and $\mathbf{U}_l$ is the pseudoinverse of the mode-$l$ expansion of $\mathcal{R}$. The fiber CUR decomposition of the tensor $\mathcal{H}$ can be represented as:
\begin{equation}
    \mathcal{H}\approx \mathcal{R}\times_{l=1}^{k}(\mathbf{C}_l\mathbf{U}_l)
    \label{eq:CUR3}
\end{equation}
where $\times_l$ represents tensor times matrix along the $l^\text{th}$ mode. The performance of the CUR decomposition is highly dependent on the selection of $\mathbf{C}$ and $\mathbf{R}$ (or $\mathcal{R}$ in high order). Key samples can result in an approximation with high accuracy. Therefore, one of the most essential tasks of CUR decomposition is to find the required samples with less computation.

\section{Method}
\label{sec:method}

\subsection{Problem Setup}
We follow the problem settings in \cite{leordeanu2005spectral}. Considering a pair of matched graphs $\mathcal{G}_1=(\mathcal{V}_1, \mathcal{E}_1)$ and $\mathcal{G}_2=(\mathcal{V}_2, \mathcal{E}_2)$, the correspondences between $\mathcal{G}_1$ and $\mathcal{G}_2$ can be represented as a binary assignment matrix $\mathbf{X}\in\{0,1\}^{n_1\times n_2}$ where $n_1=|\mathcal{V}_1|$ and $n_2=|\mathcal{V}_2|$. To solve the NP-hard graph matching problem, we denote the elements of the assignment matrix with soft constraint as $X_{ij}\in[0,1]$, where $X_{ij}$ is the probability that the $i^\text{th}$ node in $\mathcal{V}_1$ matches the $j^\text{th}$ node in $\mathcal{V}_2$. Following \cite{duchenne2011tensor}, we suppose every node in $\mathcal{G}_1$ matches exactly one node in $\mathcal{G}_2$, and every node in $\mathcal{G}_2$ matches at most one node in $\mathcal{G}_1$, i.e., $\forall i, \sum_j X_{ij}=1$ and $\forall j, \sum_i X_{ij}\leq 1$. In the rest of the paper, we call $\mathcal{G}_1$ the source graph and $\mathcal{G}_2$ target graph.\par
The compatibility matrix of second-order graph matching, $\mathbf{H}\in\mathbb{R}^{n_1n_2\times n_1n_2}$, is the unfold of the fourth-order tensor $\hat{\mathcal{H}}\in\mathbb{R}^{n_1\times n_2\times n_1\times n_2}$ where $\hat{H}_{i_1,j_1,i_2,j_2}$ represents the similarity between edges $(i_1,i_2)$ and $(j_1, j_2)$. The soft-constraint assignment matrix is flattened as $\mathbf{x}$. The second-order graph-matching problem can be represented as the optimization of the function:

\begin{equation}
    \begin{aligned}
    \max_\mathbf{x}\quad & \mathbf{x}^T\mathbf{H}\mathbf{x} \\
    s.t.\quad & \sum_{j\in\text{ind}_i} x_j=1, \forall i
\end{aligned}\label{second_order_problem}
\end{equation}
where $\text{ind}_i=\{(i-1)n_2+1,\cdots,in_2\}$ represents the index set of the $i^\text{th}$ row of $\mathbf{X}$. The compatibility matrix is symmetric, which means the compatibility between $(i_1, i_2)$ and $(j_1, j_2)$ is the same as that between $(i_2, i_1)$ and $(j_2, j_1)$. Equation \eqref{second_order_problem} can be cast into a classical Rayleigh quotient problem, and $\mathbf{x}$ is proved to be associated with the main eigenvector of $\mathbf{H}$ \cite{duchenne2011tensor}, which can be calculated with methods such as power iteration.\par 

%the optimization of the function:
%\begin{equation}
%    \begin{aligned}
%    \max_\mathbf{X}\quad & \sum_{i_1, i_2, j_1, j_2}H_{i_1,i_2,j_1,j_2}X_{i_1,j_1}X_{i_2,j_2} \\
%    s.t.\quad & \sum_j X_{ij}=1, \forall i
%\end{aligned}\label{second_order_problem}
%\end{equation}
%where the fourth-order tensor $\mathcal{H}$ represents the similarity between edges ${i_1,j_1}$ and ${i_2, j_2}$. To simplify the calculation, the tensor $\mathcal{H}$ is expanded as compatibility matrix $\mathbf{H}\in\mathbb{R}^{n_1 n_2\times n_1 n_2}$ and the assignment matrix is column-wisely expanded as vector $\mathbf{x}\in\{0,1\}^{n_1 n_2}$. The high $\mathbf{H}$ entries correspond to edges with high similarity. The compatibility matrix is symmetric, which means the similarity between $\{i_1, j_1\}$ and $\{i_2, j_2\}$ is the same as the one between $\{j_1, i_1\}$ and $\{j_2, i_2\}$. Because the problem is proven to be NP-hard, to approximate the solution of \eqref{second_order_problem}, $\mathbf{x}$ is set to obey the soft matching constraints in \cite{leordeanu2005spectral}. In this case, the problem becomes a classical Rayleigh quotient problem, and $\mathbf{x}$ is associated with the main eigenvector of $\mathbf{H}$.\par 

The $k^\text{th}$-order hypergraph-matching problem can be extended to:

\begin{equation}
    \begin{aligned}
    \max_\mathbf{x}\quad & \mathcal{H}\otimes_1\mathbf{x}\otimes_2\cdots\otimes_k\mathbf{x} \\
    s.t.\quad & \sum_{j\in\text{ind}_i} x_j=1, \forall i
\end{aligned}\label{high_order_problem}
\end{equation}

\noindent
where the $k^\text{th}$-order supersymmetric tensor $\mathcal{H}$ represents the compatibility between the hyperedges in the two graphs \cite{duchenne2011tensor}.  $\otimes_l$ is the mode-$l$ product of the tensor and vector, which is calculated as:
\begin{equation}
    (\mathcal{H}\otimes_{l}\mathbf{x})_{i_1,\cdots,i_{l-1},i_{l+1},\cdots,i_k}=\sum_{i_l}H_{i_1,\cdots,i_l,\cdots,i_k}x_{i_l}
\end{equation} 

%Hypergraph matching on a full compatibility tensor is both time and space-consuming. To reduce the exhaustive space and time complexity, \cite{duchenne2011tensor} proposed an ANN-based method to represent the compatibility tensor as a sparse tensor. It first calculates the full compatibility tensor blocks between $t$ sampled hyperedges in the source graph and all the hyperedges in the target. Then $r_1$ largest entries in each tensor block will be selected as the entries in the compatibility tensor. The time complexity of \eqref{high_order_problem} will reduced to $O(tr_1)$, which is far less than $O(n_1^kn_2^k)$. However, the calculation of each tensor block still requires at least $O(n_2^k)$ time and space complexity.

\subsection{CUR-based Second-Order Graph Matching}\label{initial_guess}

Dealing with large-scale graph matching problems, with thousands of paired nodes, is not feasible due to the terabytes of computer memory required for the compatibility matrix $\mathbf{H}$. CURSOR estimates the low-rank approximation of the compatibility matrix with CUR decomposition. Instead of directly generating the whole matrix with huge memory usage, it calculates a small number of rows and columns. Because of the symmetric property of the compatibility matrix, the column sampling is sufficient to decrease the computational complexity. By randomly selecting $c$ columns from $\mathbf{H}$ as $\mathbf{C}$, the second-order compatibility matrix can be decomposed into two smaller-sized matrices $\mathbf{C}\in\mathbb{R}^{n_1n_2\times c}$ and $\mathbf{U}_*\in\mathbb{R}^{c\times c}$. Following CUR+ \cite{xu2015cur}, $\mathbf{U}_*$ is calculated as: 
\begin{equation}
    \mathbf{U}_*=\arg\min_{\mathbf{U}}\|R_\Omega(\mathbf{H})-R_\Omega(\mathbf{C}\mathbf{U}\mathbf{C}^T)\|_F
    \label{eq:CUR}
\end{equation}
where $R_\Omega(\cdot)$ is the symbol used in the original work of CUR+ \cite{xu2015cur} to represent the matrix entries, including the randomly selected entries and all the intersection elements of $\mathbf{C}$ and $\mathbf{C}^T$. $\|\cdot\|_F$ represents the Frobenius norm of the matrix. The multiplication of $\mathbf{Hx}$ to update $\mathbf{x}$ in every iteration can be simplified as:
\begin{equation}
    \mathbf{Hx}\approx\mathbf{C}\mathbf{U}_*(\mathbf{C}^T\mathbf{x})
\end{equation}
which reduces the time and space complexity in every iteration from $O(n_1^2n_2^2)$ to $O(cn_1n_2)$. For large-scale graph matching problems, $c\ll n_1n_2$.\par 

The assignment matrix, $\mathbf{X}$, is calculated with the CUR decomposition of $\mathbf{H}$ by applying a soft-constraint second-order graph-matching algorithm, like SM \cite{leordeanu2005spectral} or RRWM \cite{cho2010reweighted}. For $i\in\{1,\cdots,n_1\}$, the $k$ entries with highest probabilities in $\mathbf{X}_{i,:}$ are found as the best $k$ match set $\mathcal{P}_i^k=\{s_{ij}\}_{j=1}^k$, where $s_{ij}$ represents the index of the $j^{\text{th}}$ highest probability in $X_{i,:}$. The detailed algorithm is shown in Algorithm~\ref{alg:kguess}. The CUR-based method may lead to lower matching accuracy with fewer sampled rows and columns if only second-order matching is applied. However, it can provide a rough result to follow with higher-order graph matching, thereby avoiding the infeasible large-scale computations.

\begin{algorithm}
\caption{CUR-based second-order graph matching}
\label{alg:kguess}
\begin{algorithmic}
\STATE \textbf{Input} Point sets $P_1, P_2$ with size $n_1, n_2$, column indices $I$, entry indices $J\supseteq I$, $k$
\STATE \textbf{Output} Best $k$ match set $\mathcal{P}^k=\text{set}(\mathcal{P}_1^k,\cdots,\mathcal{P}_{n_1}^k)$.

\STATE $\mathbf{C} \gets \mathbf{H}(:, I)$ \COMMENT{calculate columns with $P_1, P_2$}
\STATE $\mathbf{U}_* \gets \text{CUR}(\mathbf{C}, \mathbf{H}(J,J))$\COMMENT{Based on Eq.~\eqref{eq:CUR}.}

\STATE $\mathbf{x} \gets \text{Matching}(\mathbf{C}\mathbf{U}_*\mathbf{C}^T)$
\STATE $\mathbf{X}\gets \text{reshape}(\mathbf{x})$
\FOR{$i=1,\cdots,n_1$}
\STATE\COMMENT{Find $k$ entries with highest probabilities for $X(i,:)$}
\STATE $\mathcal{P}_i^k\gets\{s_{ij}\}_{j=1}^k$
\ENDFOR
\end{algorithmic}
\end{algorithm}

\subsection{Fiber-CUR-based Tensor Generation} 

The ANN-based method searches for the highest compatibilities in all target hyperedges, which is time and space-consuming for large-scale hypergraph matching. Based on randomized fiber CUR decomposition, the third-order compatibility tensor can be generated with the pairwise graph-matching result to reduce the computational cost. The second-order matching result from Algorithm~\ref{alg:kguess} is $\mathcal{P}^k$, where $k\ll n_2$. The method does not need to compare the hyperedge $(i_1,i_2,i_3)\in\mathcal{E}_1$ with all the hyperedges in the target hypergraph, but with the hyperedge set $\{(j_1,j_2,:), (j_1,:,j_3),(:,j_2,j_3)\}\in\mathcal{E}_2$ where $j_1\in\mathcal{P}_{i_1}^k$, $j_2\in\mathcal{P}_{i_2}^k$, and $j_3\in\mathcal{P}_{i_3}^k$. The $r$ highest entries from each hyperedge's calculated compatibilities in the source graph are sellected. Since each hyperedge is compared with fewer target hyperedges, $r$ is much smaller than the number of selected compatibilities $r_1$ in \cite{duchenne2011tensor}.\par 

The light blue areas in Fig.~\ref{fig:method} illustrate the comparison between the prior ANN-based tensor generation methods and the proposed fiber-CUR-based method in terms of the tensor structure. The compatibility between hyperedge $(i_1,i_2,i_3)$ in the source graph and all the hyperedges in the target graph can be represented as the tensor block $\tilde{\mathcal{H}}_{i_1,i_2,i_3}=H_{\text{ind}_{i_1},\text{ind}_{i_2},\text{ind}_{i_3}}$ (orange cubes in Fig.~\ref{fig:method}). Traditional ANN-based methods first randomly select $t$ tensor blocks, then calculate each tensor block with $n_2^3$ time and space complexity and reserve the $r_1$ highest entries in each block. In CURSOR, with the second-order graph matching result $\mathcal{P}^k$, only the corresponding fibers of each tensor block, $\mathcal{C}$ are estimated, and the $r$ highest entries are selected. Consequently, the algorithm's complexity reduces from $O(tn_2^3+tr_1)$ to $O(tk^2n_2+tr)$ where $r\ll r_1$. As only the highest compatibilities matter, there is no need to calculate the redundant $\mathbf{U}$ and $\mathcal{R}$ in Eq.~\eqref{eq:CUR3}. Compared with the original randomized fiber CUR decomposition, the compatibility tensor requires less computation. The fibers in $\mathcal{C}$ are selected based on $\mathcal{P}^k$ rather than by random sampling and are closer to the key samples. Therefore, the ground truth matching compatibility can be located with a higher probability, resulting in higher matching accuracy. %We define the fiber $\mathbf{v}$ located at $(i_1,\cdots,i_k)$ of the $k^{\text{th}}$-order tensor $\mathcal{H}$ along mode-$l$ as: 
%\begin{equation}
%    \mathbf{v}_{i_1,\cdots,i_{l-1},i_{l+1},\cdots,i_k}^l=H_{i_1,\cdots,i_{l-1},:,i_{l+1},\cdots,i_k}
%\end{equation} 
The total algorithm is given as Algorithm~\ref{alg:tensor}. 

\begin{algorithm}
\caption{Fiber-CUR-based tensor generation}
\label{alg:tensor}
\begin{algorithmic}
\STATE \textbf{Input} Point sets $P_1, P_2$ with size $n_1, n_2$, best $k$ initial guess set $\mathcal{P}^k=\text{set}(\mathcal{P}_1^k,\cdots,\mathcal{P}_{n_1}^k)$
\STATE \textbf{Output} Sparse tensor $\mathcal{H}$

\STATE $\mathcal{H} \gets \text{empty tensor}$
\STATE $\mathcal{I} \gets \text{hyperedges randomly sampled from } P_1$
\FOR{$i=(i_1,i_2,i_3)\in\mathcal{I}$}
\STATE $e\gets \text{computeHyperedgeFeature}(i,P_1)$
\STATE $\mathcal{F}\gets\text{empty feature set}$
\STATE \COMMENT{Calculate corresponding fibers in all directions.}
\STATE $\mathcal{J}_1\gets\{(1:n_2, j_2, j_3)\}, j_2\in\mathcal{P}_{i_2}^k, j_3\in\mathcal{P}_{i_3}^k$
\STATE $\mathcal{J}_2\gets\{(j_1,1:n_2,j_3)\}, j_1\in\mathcal{P}_{i_1}^k, j_3\in\mathcal{P}_{i_3}^k$
\STATE $\mathcal{J}_3\gets\{(j_1,j_2,1:n_2))\}, j_1\in\mathcal{P}_{i_1}^k, j_2\in\mathcal{P}_{i_2}^k$

%\STATE $\mathcal{T}_1=\{\mathcal{P}_{s_1}, \mathcal{P}_{s_2}, :\}$
%\STATE $\mathcal{T}_2=\{:,\mathcal{P}_{s_2}, \mathcal{P}_{s_3}\}$
%\STATE $\mathcal{T}_3=\{\mathcal{P}_{s_1}, :, \mathcal{P}_{s_3}\}$
\STATE $\mathcal{J} = \text{set}(\mathcal{J}_1, \mathcal{J}_2, \mathcal{J}_3)$
\FOR{$j\in\mathcal{J}$}
\STATE $f \gets \text{computeHyperedgeFeature}(j, P_2)$

\STATE $\mathcal{F} \gets \mathcal{F}\bigcup f$
\ENDFOR
\STATE $\mathcal{S} \gets \text{search for $r$ highest similarities}(\mathcal{F}, e)$
\FOR{$s\in\mathcal{S}$}
\STATE $\text{ind}(i,j_s)\gets \text{index of $\mathcal{H}$}(i, \text{index}(s))$
\STATE $\mathcal{H}(\text{ind}(i,j_s))\gets\text{compatibility}(e,s)$
\ENDFOR
\ENDFOR
\end{algorithmic}
\end{algorithm}

\subsection{PRL-based Matching Algorithm}\label{PRL_sec}

To accelerate the convergence of the matching process, a fast hypergraph matching algorithm was developed based on probabilistic relaxation labeling (PRL) that takes advantage of the high sparsity of the compatibility tensor. The original PRL algorithm \cite{hummel1983foundations} updates the probability that a label is assigned to an object with a set of specified compatibilities, and has been applied in several graph matching problems \cite{wu2012prl,min2013flexible,chen2015probabilistic}. For each $i\in\mathcal{G}_1$ and $j\in\mathcal{G}_2$, $p_i(j)$ represents the probability that $i$ is associated with $j$, which can be updated according to the following equation:
\begin{equation}
    p_i^{(k+1)}(j)=\frac{p_i^{(k)}(j)[1+q_i^{(k)}(j)]}{\sum_mp_i^{(k)}(m)[1+q_i^{(k)}(m)]}
    \label{PRL}
\end{equation}
with 
\begin{equation}
    q_i^{(k)}(j)=\sum_{l=1}^{n_1}d_{il}[\sum_{m=1}^{n_2}r_{il}(j,m)p_l^{(k)}(m)]
\end{equation}
where $d_{il}$ is a weight factor representing the influence of $l$ on $i$ and $\sum_ld_{il}=1$. The factor $r_{il}(j,m)\in[-1,1]$ denotes the relationship between the pairwise feature of edge $(i,l)\in\mathcal{E}_1$ and $(j,m)\in\mathcal{E}_2$.\par 

In our work, define $R_{il}(j,m)=0.5(r_{il}(j,m)+1)\in[0,1]$ and set $d_{il} = n_1^{-1}$. By replacing the weighting factor $p_i^{(k)}(j)$ in Eq.~\eqref{PRL} with updated probabilities, it gives
\begin{equation}
    p_i^{(k+1)}(j)=\frac{[\sum_l\sum_mR_{il}(j,m)p_l^{(k)}(m)]^2}{\sum_i[\sum_l\sum_mR_{il}(j,m)p_l^{(k)}(m)]^2}
    \label{PRL2}
\end{equation}
which contributes to a more consistent result and faster convergence. Define
\begin{equation}
     R_{i_1i_2}(j_1,j_2)=
    \begin{cases}
        M_{i_1,j_1} & \text{$i_1=i_2$ and $j_1=j_2$} \\
        \hat{H}_{i_1,j_1,i_2,j_2} & \text{otherwise}
    \end{cases}
\end{equation}
where $\hat{H}_{i_1,j_1,i_2,j_2}$ is the corresponding value in the compatibility matrix $\mathbf{H}$ and $M_{i_1,j_1}$ is related to the first-order compatibility between node $i_1\in\mathcal{G}_1$ and $j_1\in\mathcal{G}_2$. Since $\hat{H}_{i,j,i,j}=0$ for all $i,j$, the numerator of Eq.~\eqref{PRL2} can be updated as:
\begin{equation}
     \hat{p}_i^{(k+1)}(j)= (M_{i, j} p_i^{(k)}(j)+\sum_{l}\sum_m\hat{H}_{i,j,l,m}p_l^{(k)}(m))^2
     \label{PRL4}
\end{equation}
Denote $\delta$ as $\sum_{l}\sum_m\hat{H}_{i,j,l,m}p_l^{(k)}(m)$. To increase the corresponding true matching $p_{i_0}^{(k)}(j_0)$ consistently during every iteration, $\hat{p}_{i_0}^{(k+1)}(j_0)$ must satisfy $(M_{i_0,j_0}p_{i_0}^{(k)}(j_0)+\delta)^2\geq p_{i_0}^{(k)}(j_0)$, which leads to 
\begin{equation}
    \delta\geq 0.25M_{i_0,j_0}^{-1}   
    \label{PRL5}
\end{equation}
If Eq.~\eqref{PRL5} is guaranteed, the probabilities can converge fast. The detailed derivation of the PRL-based method can be found in the supplement.\par

To extend the algorithm to third-order hypergraph matching, replace the probability set $\mathbf{p}=\{p_i(j)\}$ with the vector $\mathbf{x}$, the column-wise flattening of soft-constraint assignment matrix $\mathbf{X}$. After normalization, rewrite Eq.~\eqref{PRL4} as
\begin{equation}
    \hat{\mathbf{x}}^{(k+1)}= (\alpha\hat{\mathbf{m}}\odot\mathbf{x}^{(k)} + (1-\alpha)\mathcal{H}\otimes_1\mathbf{x}^{(k)}\otimes_2\mathbf{x}^{(k)})^2
    \label{PRL_H}
\end{equation}
where $\odot$ is the element-wise multiplication and $\alpha\in [0,1]$ is a balance weight between the first and third-order compatibilities. The square calculation in Eq.~\eqref{PRL_H} is also element-wise. The first-order compatibility vector $\hat{\mathbf{m}}$ is obtained by column-wise flattening $\hat{\mathbf{M}}$. The steps are shown in Algorithm~\ref{alg:PRL}. For a tensor with high sparsity, the non-zero entries of $\mathcal{H}\otimes_1\mathbf{x}^{(k)}\otimes_2\mathbf{x}^{(k)})^2$ are concentrated. Therefore $\mathbf{x}^{(k)}$ has a fast convergence speed if the ground truth compatibility entries are selected in the sparse tensor. According to Eq.~\eqref{PRL5}, a lower $\alpha$ can be set for the tensor with high sparsity to achieve faster convergence.

\begin{algorithm}
\caption{PRL-based hypergraph matching}
\label{alg:PRL}
\begin{algorithmic}
\STATE \textbf{Input} Sparse compatibility tensor $\mathcal{H}$, initial assignment matrix $\mathbf{X}$, first-order compatibility vector $\hat{\mathbf{m}}$, $\alpha$
\STATE \textbf{Output} Soft-constraint assignment matrix $\mathbf{X}$
\REPEAT
\STATE $\mathbf{x} \gets \text{flatten}(\mathbf{X})$
\STATE $\mathbf{\delta} \gets \mathcal{H}\otimes_1\mathbf{x}\otimes_2\mathbf{x}$
\STATE $\mathbf{x} \gets \alpha\hat{\mathbf{m}}\odot\mathbf{x}+(1-\alpha)\mathbf{\delta}$
\STATE $\mathbf{x} \gets \mathbf{x}\odot\mathbf{x}$ 
\STATE $\mathbf{X} \gets \text{reshape}(\mathbf{x})$
\STATE $\mathbf{X} \gets \text{norm}(\mathbf{X})$ \COMMENT{Normalize across columns}
\UNTIL{$\mathbf{X}$ converges}
\end{algorithmic}
\end{algorithm}

\section{Experiments}\label{experiment}

\begin{table*}

\begin{center}
%\begin{tabular}{cccc|ccc|ccccc}
\begin{tabular}{rrrr|rrr|rrrrr}
\hline
&&&&\multicolumn{3}{c}{ADGM}&\multicolumn{5}{|c}{CURSOR}\\
\hline
$n_1$&$n_2$&$\sigma$&$t$&$r_1$&Memory ($\mathcal{H}$)&\textbf{Accuracy}&$c$&$k$&$r$&Memory ($\mathbf{H}+\mathcal{H}$)&\textbf{Accuracy}\\
\hline
30 & 30&0.02&900&900&89.35MB&1 & 15 & 5 &5&0.62MB& 1\\
30 & 50&0.02&1500&2500&413.88MB&1 & 15& 5 &5&0.82MB& 1\\
50& 50&0.02&2500&2500&701.80MB&1 & 20 & 7 &7&2.37MB& 1\\
50& 100&0.02&5000&10000&5.45GB&0.974 & 100 & 10 &10&9.62MB& 0.982\\
100& 100&0.02&10000&10000&11.28GB&1 & 100 & 15 &20&30.65MB& 1\\
300& 300&0.01&30000&-$^*$&-& -& 200 & 20 &20&215.92MB& 1\\
500& 500&0.01&50000&-&-&- & 300 & 25 &30&783.85MB& 0.969\\
800& 800&0.005&80000&-&-&-&400&30&50&2.02GB&0.973\\
1000& 1000&0.005&100000&-&-&-&500&50&80&5.03GB&0.992\\

\hline
\end{tabular}
\caption{Results on the synthetic dataset with ADGM and CURSOR.}
\label{tab-data}
\end{center}
\footnotesize{$^*$System runs out of memory.}
\end{table*}
CURSOR was compared with four learning-free third-order ANN-based hypergraph matching algorithms: Tensor Matching (TM) \cite{duchenne2011tensor}, Hypergraph Matching via Reweighted Random Walks (RRWHM) \cite{lee2011hyper}, BCAGM in third-order (BCAGM3) \cite{nguyen2015flexible}, and Alternating Direction Graph Matching (ADGM)\cite{le2017alternating}. The experiments were conducted on the original implementations provided by their authors. The proposed tensor generation method was integrated into each of these algorithms, represented as CURSOR+TM/RRWHM/BCAGM3/ADGM in the experiments. The all-ones vector was set as the starting point, and the Hungarian algorithm \cite{kuhn1955hungarian} turned the output into a proper matching. CURSOR was also compared with the state-of-the-art deep-learning-based algorithm HNN-HM \cite{liao2021hypergraph} on the House and Hotel dataset, which is relatively small-scale. Since HNN-HM failed for datasets with $n_1>40$ under the same memory constraint, it was not compared with CURSOR on other datasets. The hyperparameter $\alpha$ in Algorithm~\ref{alg:PRL} was set to 0.2 during the experiments. The experiments were run on a computer with an Intel Core i7-9700 CPU @ 3.00 GHz and 16 GB of memory. All quantitative results were obtained by 50 trials. Due to space limitations, the ablation studies and parameter sensitivity analysis are given in the supplement.

The compatibility features for each order and the parameters for the experiments were set as:\par 
\noindent\textbf{First-order compatibility.} The first-order compatibility matrix $\mathbf{M}$ in Eq.~\eqref{PRL_H} for all experiments was calculated as:
\begin{equation}
    M_{i,j}=\exp(-\gamma_0\|f_i-f_j\|_2)
\end{equation}
where $f_i$ and $f_j$ are the normalized coordinates of $i^{th}$ point in $P_1$ and $j^{th}$ point in $P_2$, respectively. The coordinates are normalized by subtracting the mean value of the coordinates in each set. $\gamma_0$ is the inverse of the mean value of all the distances from points in $P_1$ to the ones in $P_2$. $\mathbf{M}$ is then flattened column-wise as the first-order compatibility vector $\hat{\mathbf{m}}$.\par
\noindent\textbf{Second-order compatibility matrix.} The pairwise compatibility feature calculation of CURSOR followed \cite{khan2019image}, which found a balance between rotation and scale invariance. $c$ columns, as $\mathbf{C}$, and another $3c^2$ entries of $\mathbf{H}$ were randomly selected with a uniform distribution for the CUR decomposition. The soft-constraint assignment matrix was computed with the second-order PRL-based algorithm presented in Eq.~\eqref{PRL4}.\par
\noindent\textbf{Third-order compatibility tensor.} Following~\cite{duchenne2011tensor}, the same third-order compatibility feature calculation for the ANN-based and CURSOR methods was applied. The same $t$ randomly selected hyperedges in the source graph were used for all the methods. The ANN-based and CURSOR methods generated the tensor with $r_1$ and $r$ highest compatibilities from the target graph, respectively.\par

\subsection{Large-Scale Random Synthetic Dataset}\label{synthetic}

One thousand two-dimensional points, $P$, were sampled from a Gaussian distribution $\mathcal{N}(0,1)$. Then, Gaussian deformation noise $\mathcal{N}(0, \sigma^2)$ was added to $P$ as point set $Q$. During the experiments, $n_1$ points from $P$ were selected as source graph $\mathcal{G}_1$, and $n_2$ points containing the corresponding matching of $\mathcal{G}_1$ and outliers from $Q$ were chosen as target graph $\mathcal{G}_2$.\par

CURSOR was evaluated on synthetic datasets with increasing problem scales. Results of the ADGM algorithm based on ANN, whose accuracy was the highest among all the prior state-of-the-art hypergraph matching algorithms, were given. The parameter settings of ADGM, including $t$ and $r_1$, strictly followed the original work \cite{le2017alternating}. CURSOR can deal with large-scale scenarios and achieve privilege-matching accuracy with much less memory usage (Table~\ref{tab-data}). CURSOR was capable of solving 1000-vs-1000 matching problems with high accuracy, while ADGM failed to generate the tensor when $n_1>100$ under the same memory constraint. A more detailed memory footprint analysis and the potential bottleneck of CURSOR will be provided in the supplement.\par

\subsection{Templates with Specific Shapes}

%We further implement the algorithms on a smaller dataset for comparison. We first set $\sigma$ as 0.02, $n_1=100$, and vary $n_2$ from 100 to 300, which means there are outliers from 0 to 200 in the target graph. The results in Fig.~\ref{fig:syn}~\subref{subfig:syn_nnz} and Fig.~\ref{fig:syn}~\subref{subfig:syn_acc} show that with about ten times sparser than the compatibility tensor using the ANN-based method, our CUR-based method can generate tensors leading to much higher accuracy. Then we set $n_2$ as 100 with no outlier and vary $\sigma$ from 0 to 0.1. The result in Fig.~\ref{fig:syn}~\subref{subfig:syn_nacc} shows that the matching algorithms with the CUR-based method have higher matching accuracy than the ones with the ANN-based method. The CUR-based method can generate compatibility tensors with higher robustness.\par 
\begin{figure*}[htbp]
\centering
\subfloat[Rotation]{\includegraphics[width=0.23\textwidth]{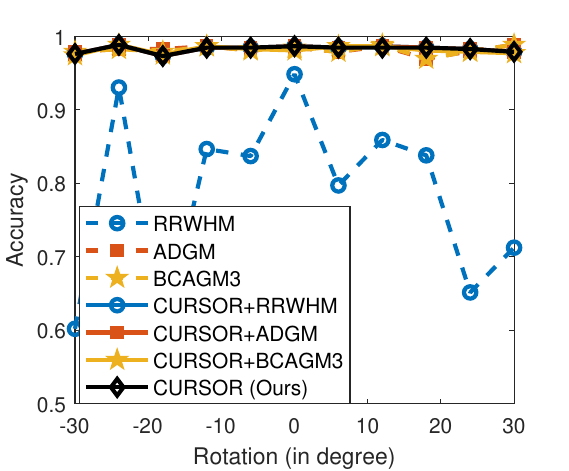}\label{rotation_acc}}
 \subfloat[Scaling]{\includegraphics[width=0.23\textwidth]{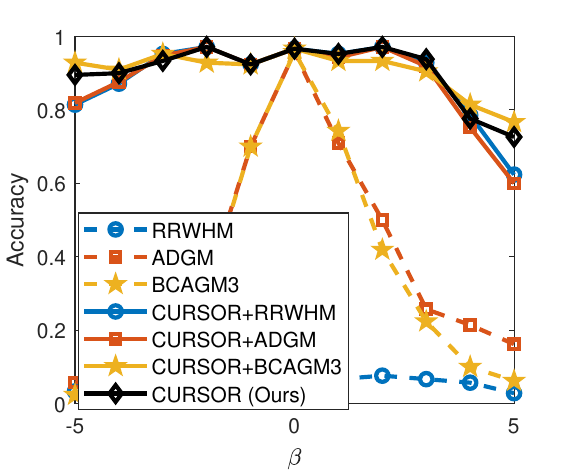}\label{scale_acc}}
 \subfloat[Adding Noise]{\includegraphics[width=0.23\textwidth]{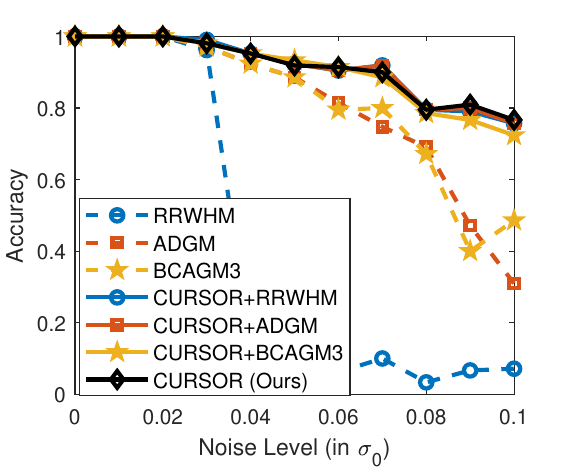}\label{noise_acc}}
 \subfloat[Adding Outliers]{\includegraphics[width=0.23\textwidth]{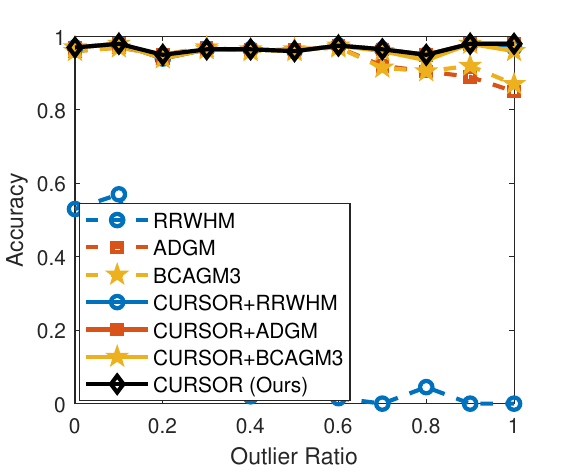}\label{outlier_acc}}
 
\caption{Matching result with deformation \subref{rotation_acc} rotation with angle $[-30^\circ, 30^\circ]$, \subref{scale_acc} scale on $x$-coordinate with scale factor $1.1^\beta$ where $\beta\in[-5,5]$, \subref{noise_acc} adding noise with $\sigma/\sigma_0=[0,0.1]$, and \subref{outlier_acc} adding $n_l$ outliers where the outlier ratio$=n_l/n_1$.}

\label{fig:shape_acc}
\end{figure*}

\begin{figure*}[htbp]
\centering
\subfloat[House: 20 pts vs 30 pts]{\includegraphics[width=0.23\textwidth]{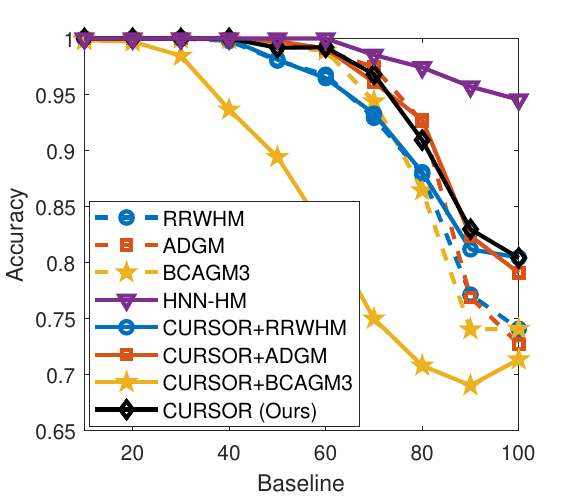}}
 \subfloat[House: 30 pts vs 30 pts]{\includegraphics[width=0.23\textwidth]{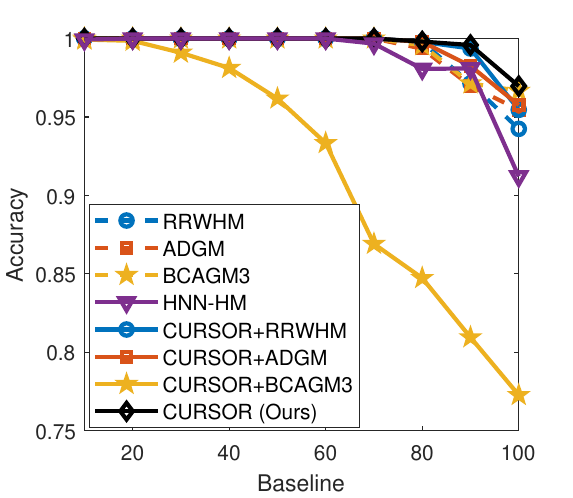}}
 \subfloat[Hotel: 20 pts vs 30 pts]{\includegraphics[width=0.23\textwidth]{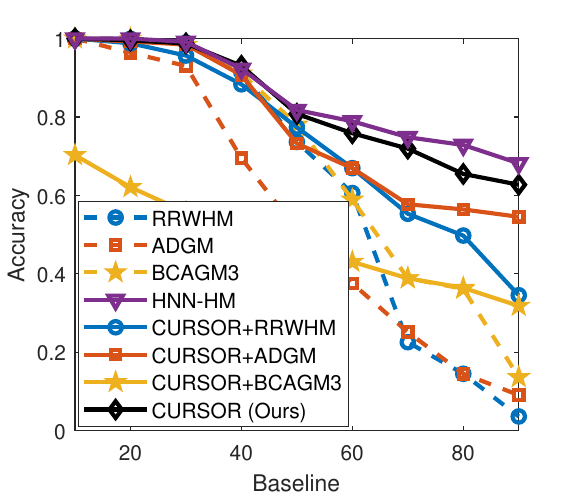}}
 \subfloat[Hotel: 30 pts vs 30 pts]{\includegraphics[width=0.23\textwidth]{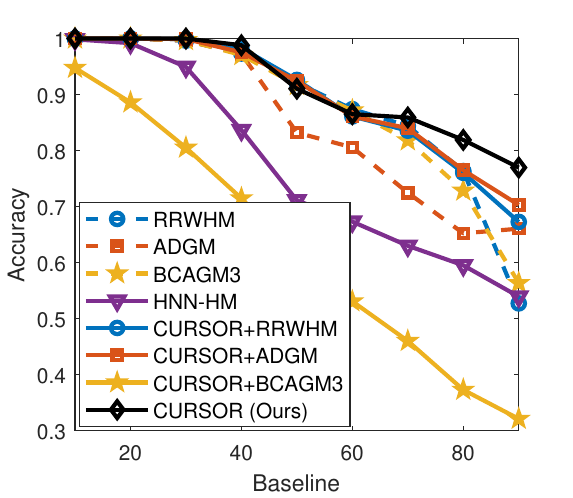}}
\caption{Comparison results on the House and Hotel dataset with various matching algorithms. The dashed curves represent the matching results on the compatibility tensors using ANN. The solid curves with the same color denote the matching accuracy on tensors generated by CURSOR with the same hypergraph matching algorithms.}
\label{fig:house}
\end{figure*}

To analyze the robustness of CURSOR to different shape deformations, middle-scale templates with specific shapes, which have been commonly used in previous works \cite{wang2019functional, khan2019image, myronenko2010point, jian2010robust} were evaluated. Four 2D templates were chosen: whale with 150 points, Chinese character with 105 points, UCF fish with 98 points, and tropical fish with 91 points. The target points were generated by rotation, scaling, with noise and outliers added. For CURSOR, $c=100$, $k=5$, $r=25$ and for ANN-based methods, $r_1=300$. For both types of methods $t=0.3n_1n_2$.\par

Different deformations were added on all the target points. For the rotation case, the source points were rotated with angle $\theta\in[-30^\circ,30^\circ]$, then noise with $\sigma=0.02\sigma_0$, where $\sigma_0$ is the standard deviation of all the source point coordinates, was added. For the scaling case, only the $x$-coordinate of the 2D point was scaled with the scaling factor $s=1.1^\beta$, where $\beta\in[-5,5]$, then noise with $\sigma=0.02\sigma_0$ was added. To evaluate variant noise level, we vary noise with $\sigma$ ranging from 0 to $0.1\sigma_0$ with step $0.01\sigma_0$. To add outliers, assume the mean value of the source points is $\mathbf{\mu}_0$ and randomly add $n_l$ outliers from a Gaussian distribution $\mathcal{N}(\mathbf{\mu}_0,\sigma_0^2)$, where the outlier ratio $n_l/n_1\in[0,1]$.\par

Due to the rotational invariance of hyperedge features, almost all the algorithms show an average accuracy of nearly 1 under a variant of rotation deformation whether ANN or CURSOR (Fig.~\ref{fig:shape_acc}). %because of the rotation invariance of hyperedge features. 
The curves of other three cases demonstrate that the matching algorithms with CURSOR outperform ANN. For example, with scaling, the hyperedge features were substantially damaged because the scaling process only scaled the $x$-coordinate of the target points. 
%The ANN-based method selects the nearest neighbors in the whole $n_2^3$ tensor block, which increases the probability of selecting the wrong neighbors if the features are heavily damaged. 
As the scale factor increased, the matching results of the ANN-based algorithms significantly decreased. In contrast, CURSOR generated compatibility tensors with higher robustness. The performance of the ANN matching algorithms was significantly improved and became competent after integrating the CURSOR tensor generation method. RRWHM performed much worse than other algorithms in all the cases but with the assistance of the compatibility tensor generated by CURSOR, its results became stable and showed comparable results to the other algorithms.

\subsection{CMU House and Hotel Dataset}\label{house}

Previous works used the CMU House and Hotel datasets to evaluate the matching algorithms \cite{duchenne2011tensor, lee2011hyper, le2017alternating,liao2021hypergraph, nguyen2015flexible}. These datasets have 30 manually labeled feature points on a rotating 3D house or hotel model tracked over 111 and 101-frame image sequences, respectively. The experiment settings of \cite{le2017alternating} were followed by matching all possible pairs with baseline (the separation between frames) varying from 10 to 100 in intervals of 10 for the House dataset and from 10 to 90 for the Hotel dataset. $n_1$ points were randomly selected in the first image to match all the points in the second for both datasets, where $n_1$ equaled 20 and 30 as two separate experiments. $t=n_1n_2$ tuples were selected from the first image. For ANN-based methods, $r_1=200$ nearest neighbors were chosen for each hyperedge. For CURSOR, $r$ was set to 25 and $c=300$. For the learning-based HNN-HM, the training process of \cite{liao2021hypergraph} on the House dataset was followed and the model was validated on both datasets.\par

\begin{figure*}[htbp]
\centering
%\begin{minipage}{.23\textwidth}
%\subfloat{\includegraphics[width=1\textwidth]{figure/exp/car/nnz_car.eps}}
% \\
% \subfloat{\includegraphics[width=1\textwidth]{figure/exp/car/acc_car.eps}}
 
%\end{minipage}
\subfloat[NNZ of Tensor]{\includegraphics[width=0.19\textwidth]{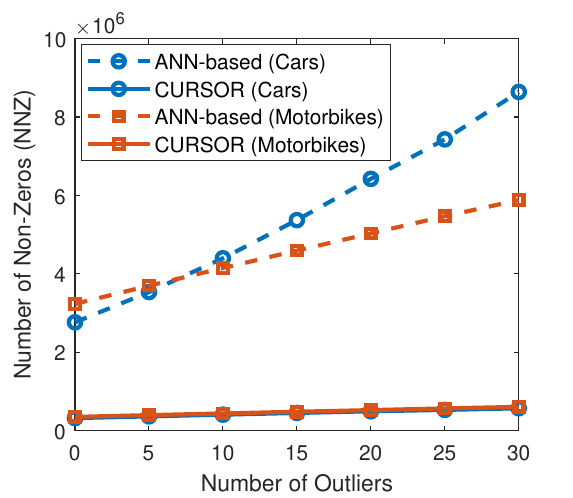}\label{subfig:nnz}}
\subfloat[Motorbikes Dataset]{\includegraphics[width=0.4\textwidth]{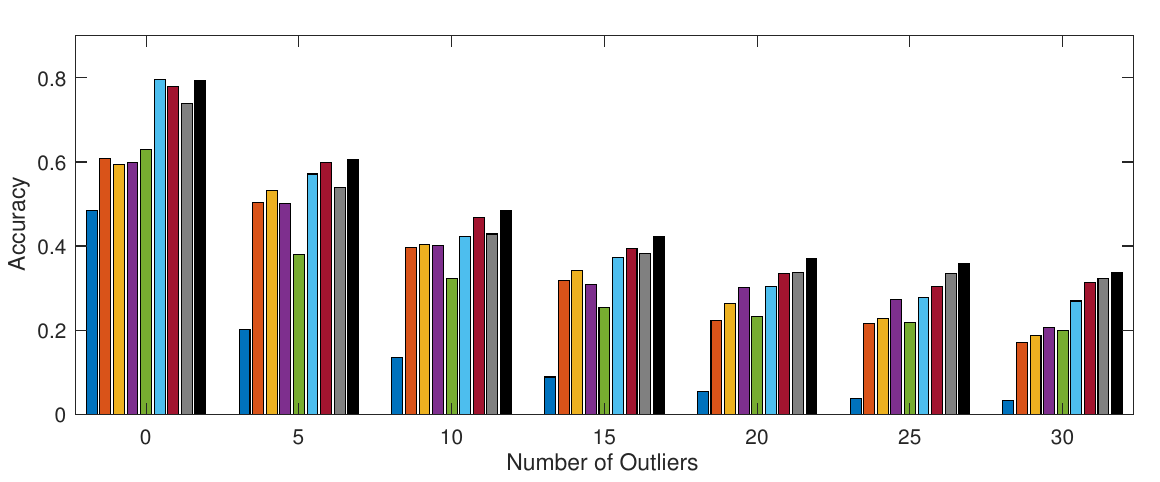}\label{subfig:motor}}
\subfloat[Cars Dataset]{\includegraphics[width=0.4\textwidth]{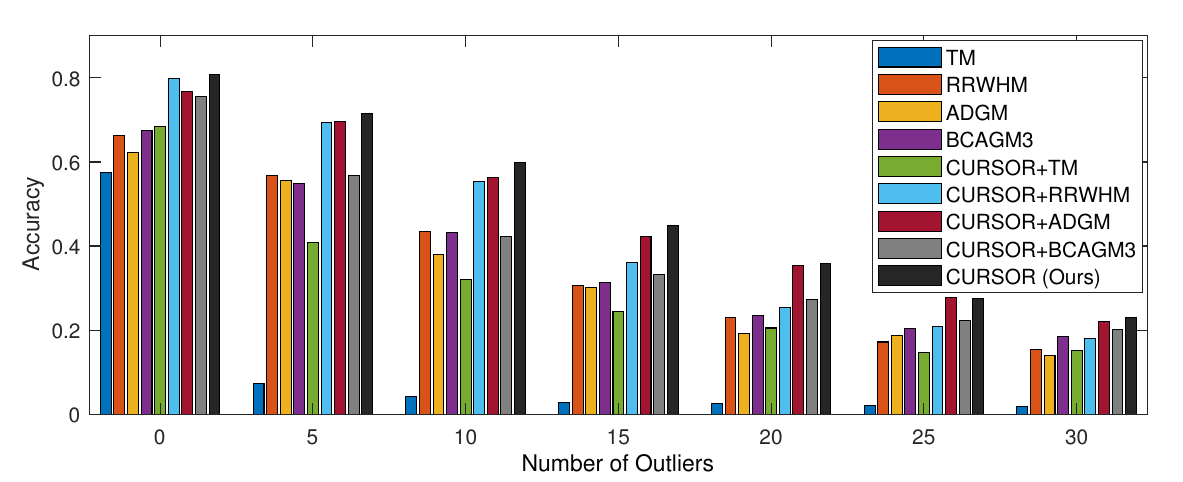}\label{subfig:car}}

%\begin{minipage}{.23\textwidth}
%    \subfloat{\includegraphics[width=1\textwidth]{figure/exp/motor/nnz_motor.eps}}\\
%    \subfloat{\includegraphics[width=1\textwidth]{figure/exp/motor/acc_motor.eps}}
%\end{minipage}

\caption{The Cars and Motorbikes datasets with CURSOR and state-of-the-art hypergraph matching algorithms. \subref{subfig:nnz} The number of non-zero compatibilities with ANN-based methods and CURSOR. The matching accuracy on the \subref{subfig:motor} Motorbikes and \subref{subfig:car} Cars datasets.}
\label{fig:cars}
\end{figure*}

\begin{figure*}[htbp]
\centering
\subfloat[30 pts vs 40 pts (10 outliers)]{\includegraphics[width=0.21\textwidth]{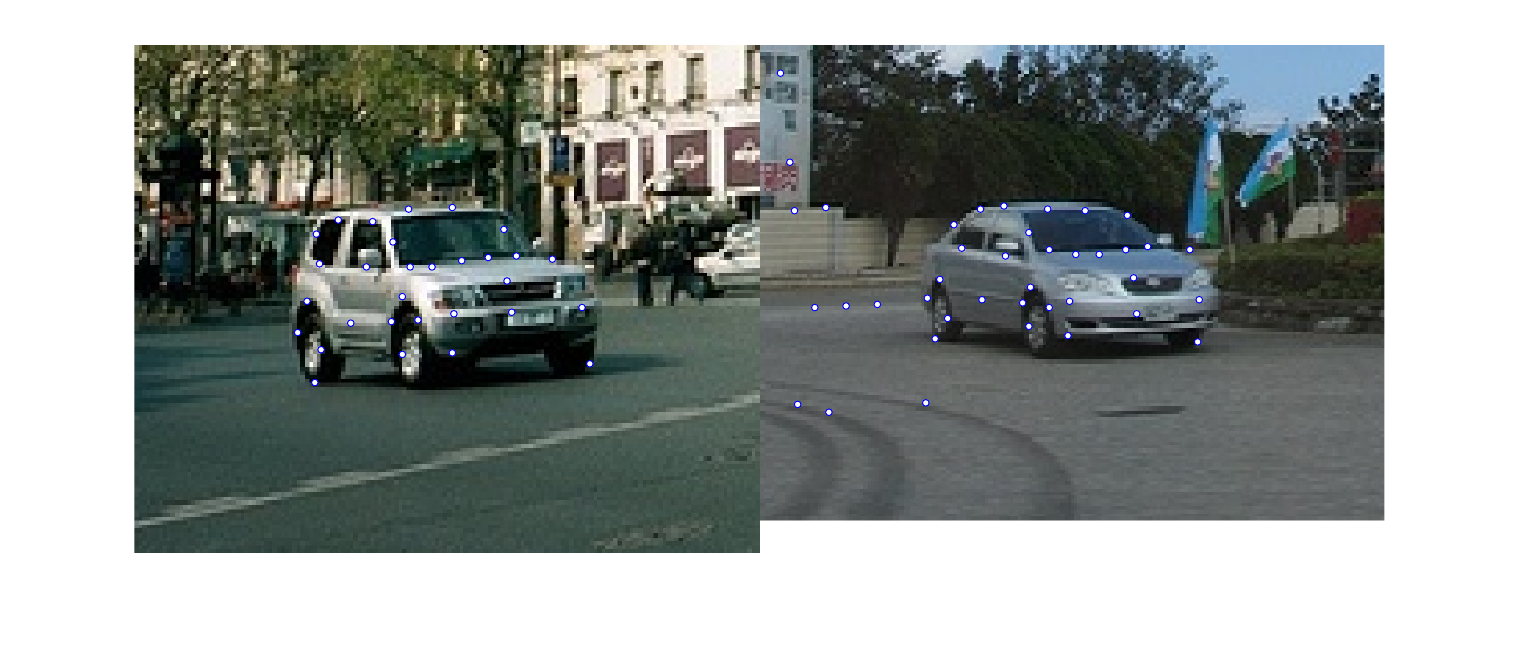}}
\subfloat[ADGM 3/30]{\includegraphics[width=0.21\textwidth]{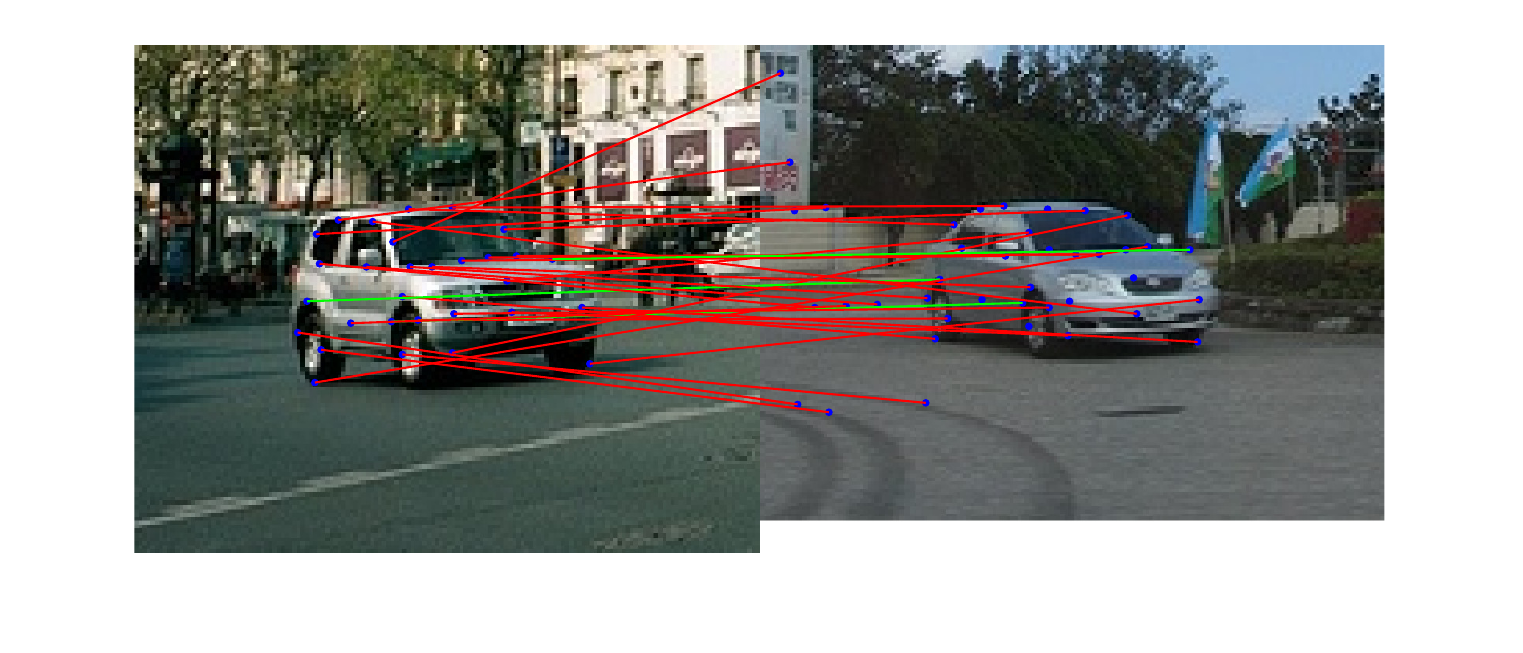}}
 \subfloat[CURSOR+ADGM 25/30]{\includegraphics[width=0.21\textwidth]{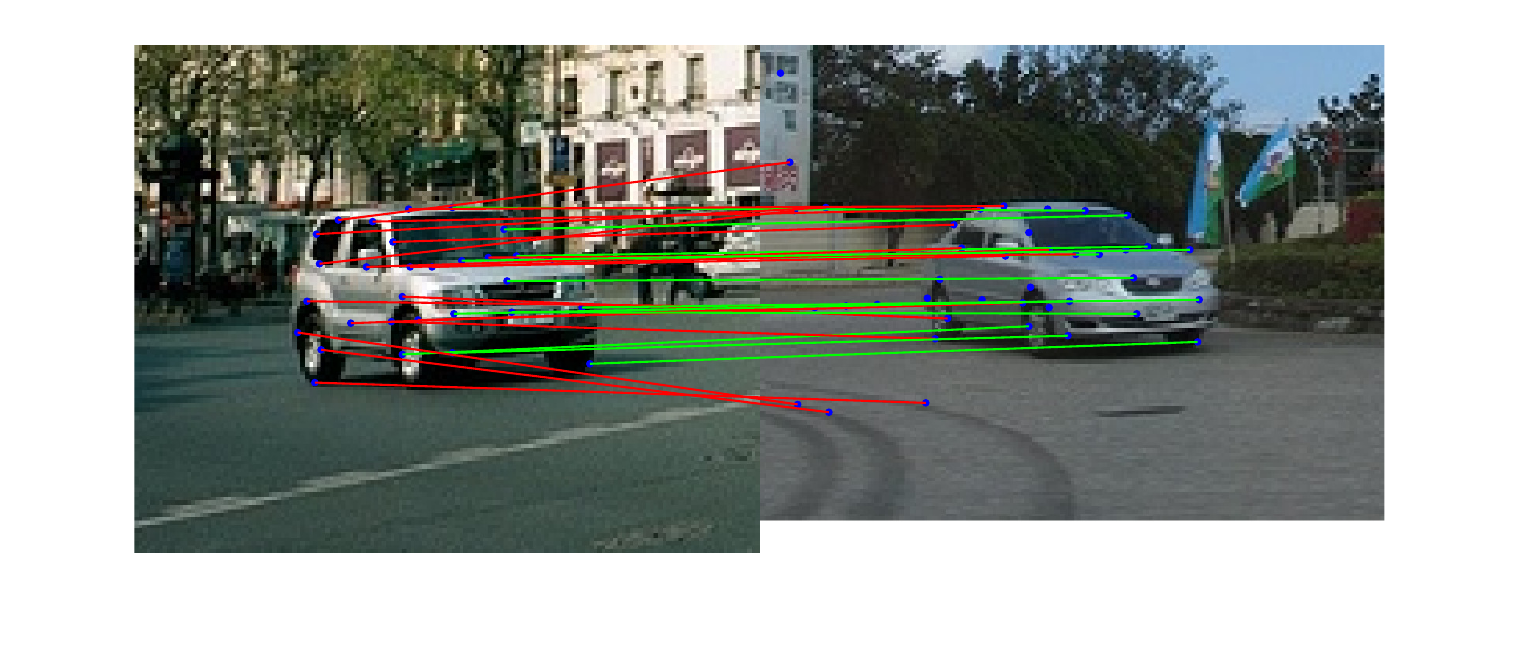}}
 \subfloat[CURSOR (Ours) 30/30]
 {\includegraphics[width=0.21\textwidth]{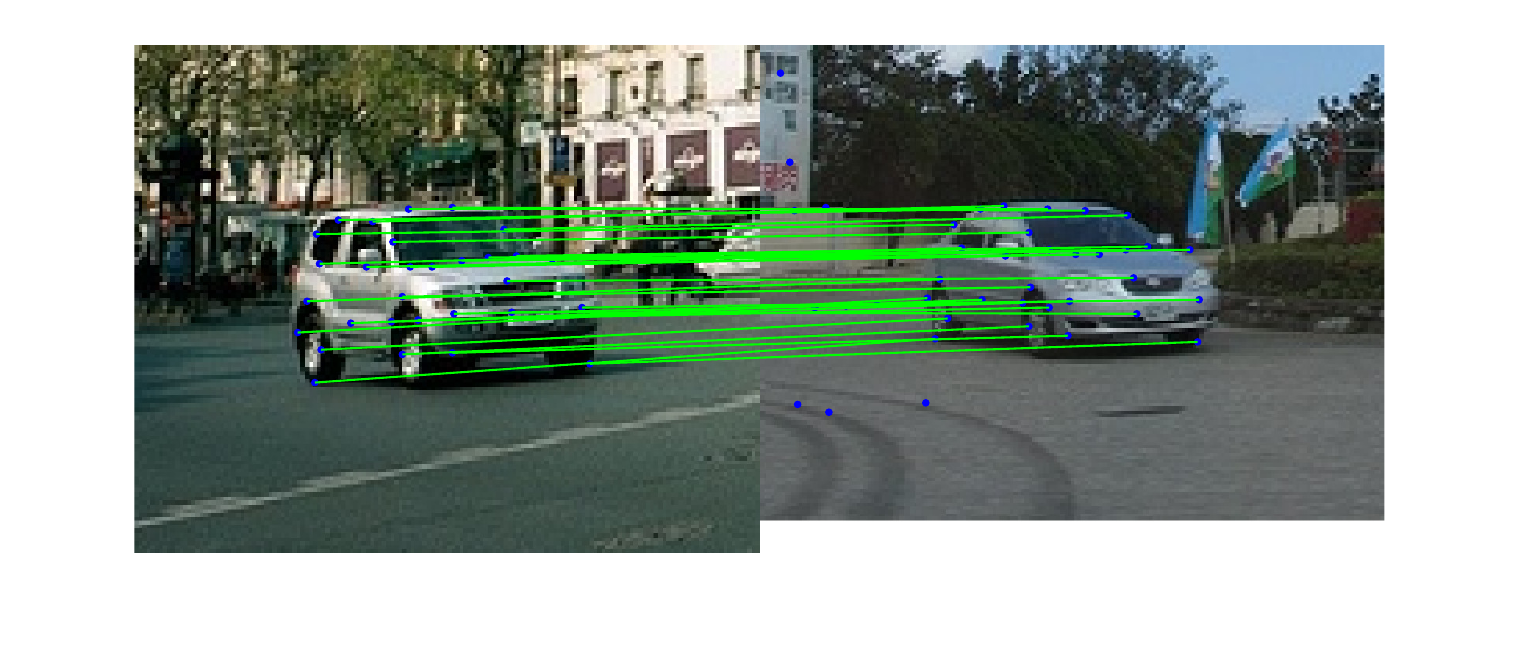}}\\
\subfloat[52 pts vs 67 pts (15 outliers)]{\includegraphics[width=0.21\textwidth]{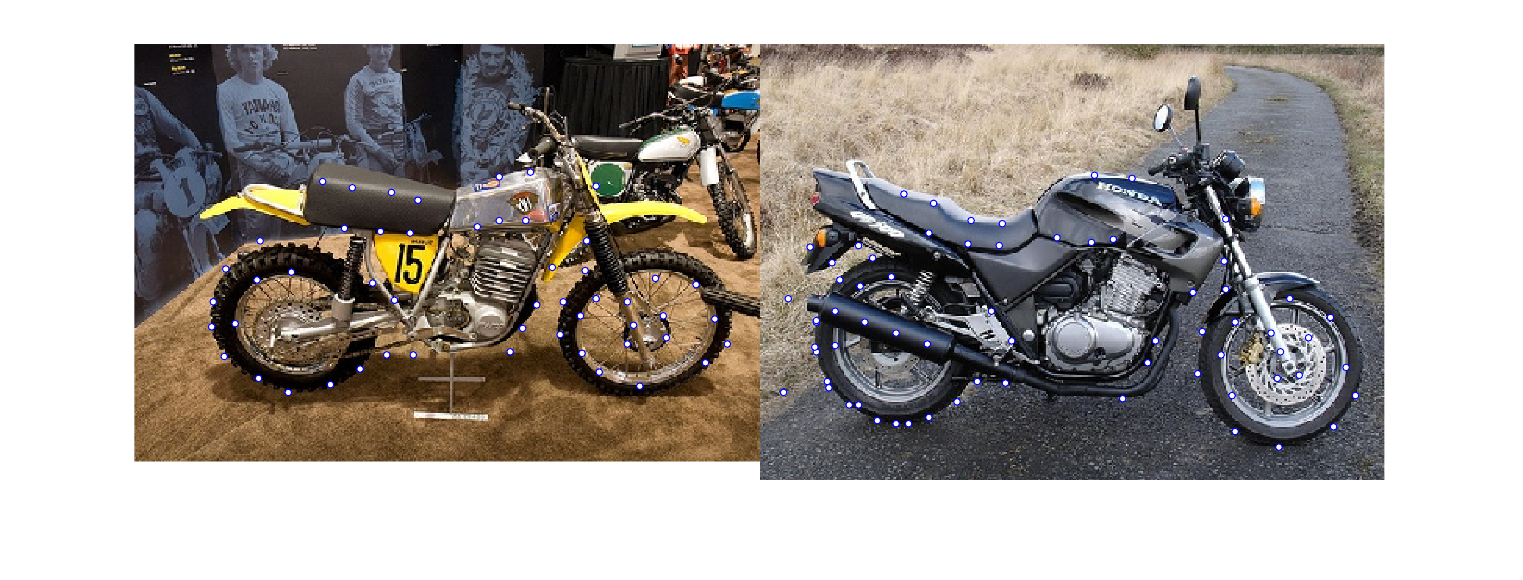}}
\subfloat[BCAGM3 36/52]{\includegraphics[width=0.21\textwidth]{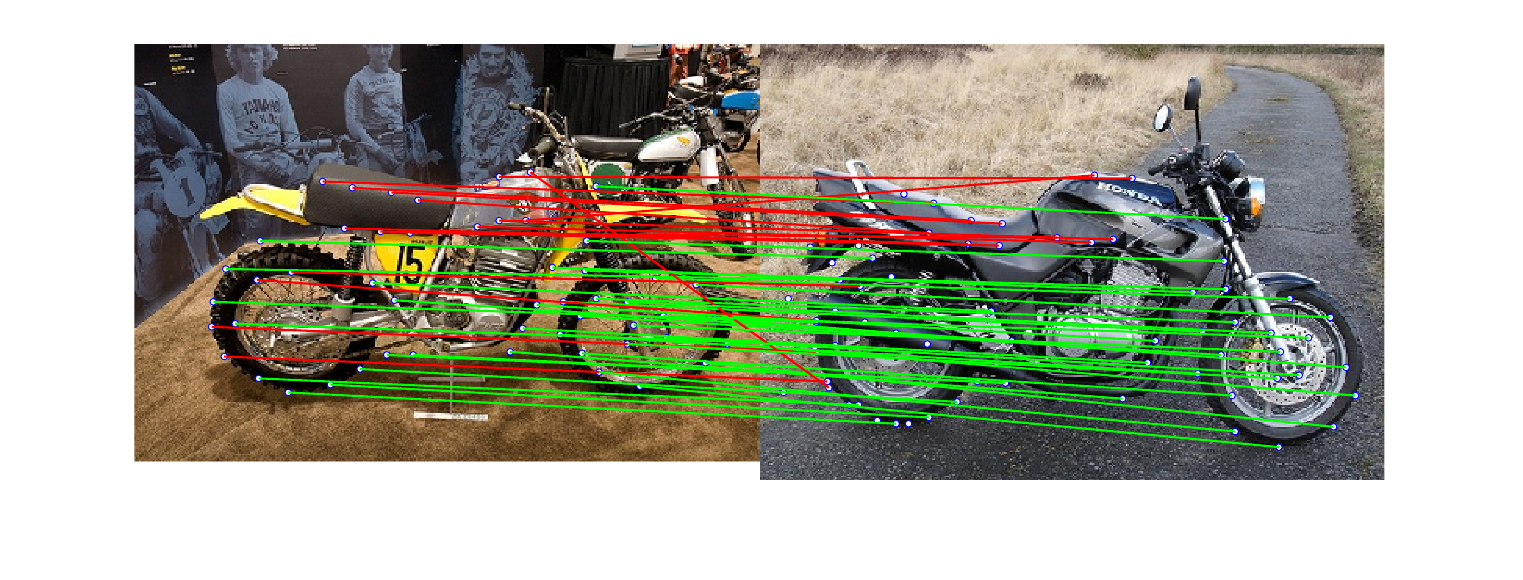}}
 \subfloat[CURSOR+BCAGM3 50/52]{\includegraphics[width=0.21\textwidth]{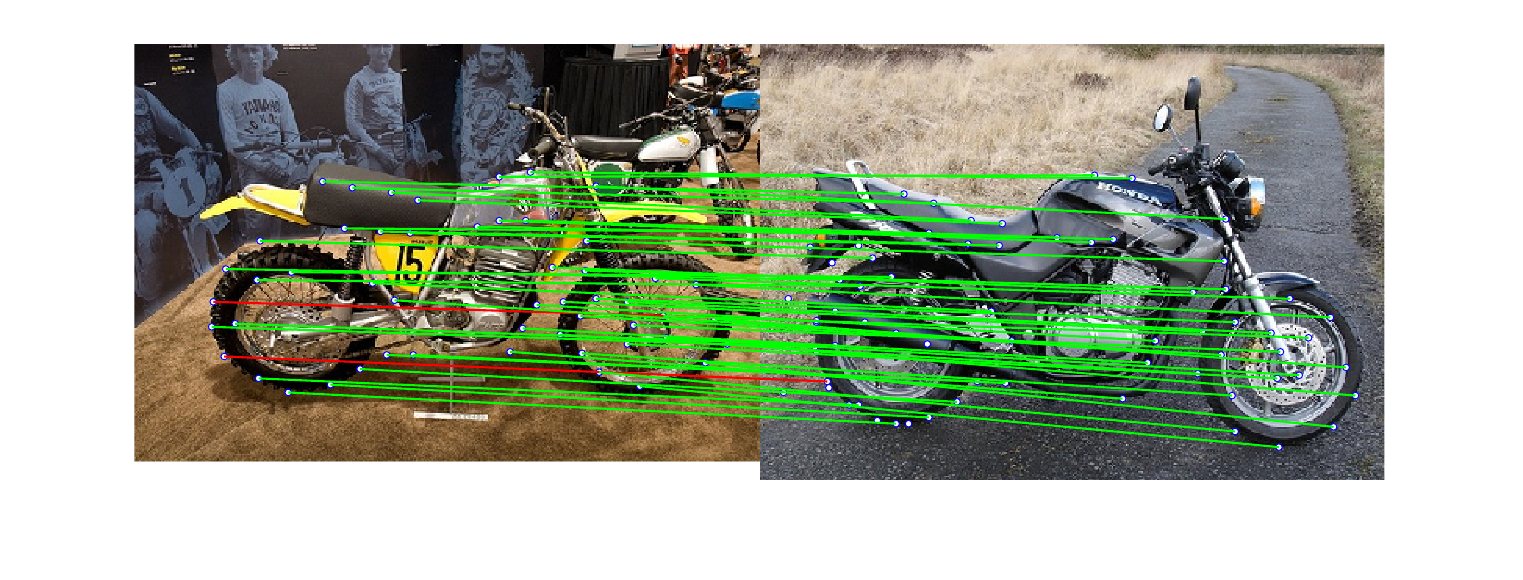}}
 \subfloat[CURSOR (Ours) 52/52]
 {\includegraphics[width=0.21\textwidth]{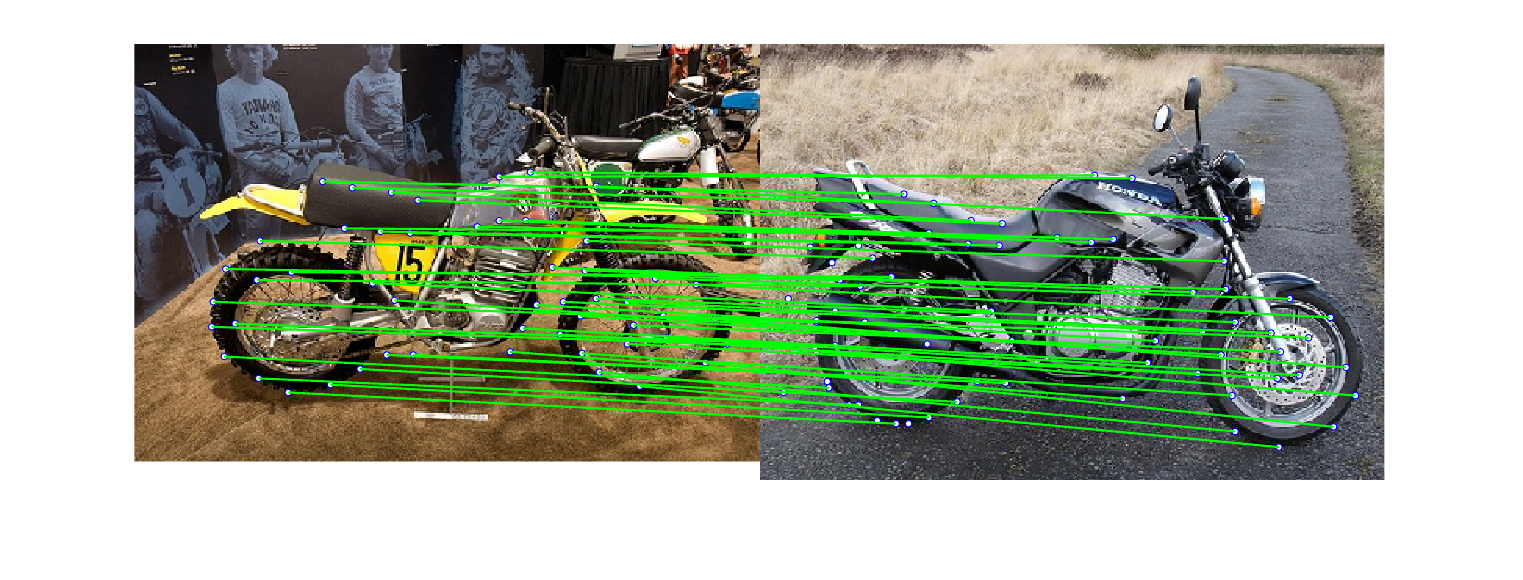}}
\caption{Car and Motorbike matching examples. Top row Car dataset, bottom row Motorbike dataset. Each example shows the matched results with the highest accuracy among trials. The green and red lines denote matches and mismatches, respectively.}
\label{fig:motor_example}
\end{figure*}

The sequence-matching results are given in Fig.~\ref{fig:house}. All the learning-free-based matching algorithms combined with CURSOR surpassed the original ANN-based ones except for the BCAGM3 algorithm. The BCAGM3 algorithm with CURSOR frequently made descent errors during the experiment as its matching accuracy heavily decreased when the compatibility tensor was too sparse. CURSOR obtained comparable matching accuracy to the state-of-the-art learning-free algorithms when the outlier number was ten and achieved the best matching accuracy under the 30-vs-30 case on the House dataset. For the Hotel dataset, CURSOR %first considers the distance and rotation invariance during the second-order graph matching process and, therefore, 
achieved a much higher matching accuracy than ANN, but was less accurate than HNN-HM on 20-vs-30 problems as HNN-HM was specifically trained on the House Dataset. %  \red{it has much higher matching accuracy than the current learning-free-based hypergraph matching algorithms on 20-vs-30 problems of the same dataset. However, it still falls behind on 30-vs-30 ones. With the same trained model for the Hotel dataset, HNN-HM has higher matching accuracy in 20-vs-30 problems but heavily falls behind in the 30-vs-30 problems.} 
It is noteworthy that CURSOR solved the matching problem with a unified model while HNN-HM must first train separating models in different datasets to achieve a better matching result. More detailed analysis will be provided in the supplement to show the effectiveness of CURSOR.\par

\subsection{Car and Motorbike Dataset}

The Car and Motorbikes dataset~\cite{leordeanu2012unsupervised} consists of 30 real-life car image pairs and 20 motorbike image pairs, and was used in previous works to evaluate matching algorithms \cite{lee2011hyper, le2017alternating, nguyen2015flexible}. In this experiment, all inlier points in both images were kept and labeled outlier points were randomly chosen in the second image, with the number varying from 0 to 30 in step of 5. Every image pair in both datasets was matched. $t=n_1n_2$ hyperedges were selected in the first image. For ANN-based methods, $r_1$ was set as $0.3n_1n_2$. CURSOR selected $r=50$ highest compatibilities in each tensor block. During experiments $k=10$ and $c=300$ .\par 

Figure~\ref{subfig:nnz} reported the average number of non-zero elements in the compatibility tensors generated by the two types of methods. The average accuracy with the ANN-based methods and CURSOR was shown in Fig.~\ref{subfig:motor} and Fig.~\ref{subfig:car}. %with the average number of non-zero elements in the tensors generated by the two types of methods shown (The comparison results are shown in Fig.~\ref{fig:cars}). 
All algorithms integrated with CURSOR consistently improved their matching performance with a compatibility tensor more than ten times sparser than the ANN-based methods. In most cases, the default PRL-based matching algorithm achieved higher accuracy. Figure~\ref{fig:motor_example} shows matching examples. Combined with CURSOR, other hypergraph matching algorithms showcased fewer mismatches. CURSOR, with the PRL-based algorithm, obtained the highest matching accuracy.

\section{Conclusion}\label{conclusion}
We propose CURSOR, a cascaded mixed-order hypergraph matching framework based on CUR decomposition for scalable graph matching. The framework contains a CUR-based second-order graph matching algorithm and a fiber-CUR-based tensor generation method, which significantly decreases the computational cost, and can be seamlessly integrated into existing state-of-the-art hypergraph matching algorithms to enhance their performance. A PRL-based hypergraph matching algorithm for sparse compatibility tensors is developed to accelerate the convergence. Experiment results demonstrated that CURSOR contributes to a higher matching accuracy with a sparser tensor, which has more potential utility in the big-data era. In future work, we plan to develop a more principled adaptive scheme to optimize the parameters of CURSOR so that it can perform better with fewer computations on larger-scale tasks. \par 

%Although the proposed method efficiently increases the matching accuracy, the parameters are still manually set in different datasets. In future work, we plan to optimize the method further for a more principled adaptive scheme to different parameters in the algorithm so that it can perform better with fewer computations and larger-scale problems than the current method.

\section*{Acknowledgment}
This work is supported by the Hong Kong Innovation and Technology Commission (InnoHK Project CIMDA), the Hong Kong Research Grants Council (Project 11204821), and City University of Hong Kong (Project 9610034).
%\input{sec/fake_file}
%\nolinenumbers
{
    \small
    \bibliographystyle{ieeenat_fullname}
    \bibliography{main}
}

\clearpage

\maketitlesupplementary

%\linenumbers
\section{PRL-based Method Derivation}

Consider source graph $\mathcal{G}_1=\{\mathcal{V}_1,\mathcal{E}_1\}$ and target graph $\mathcal{G}_2=\{\mathcal{V}_2,\mathcal{E}_2\}$. Since the compatibility matrices/tensors are non-negative in graph matching, the original compatibility coefficient between $(i,l)\in\mathcal{E}_1$ and $(j,m)\in\mathcal{E}_2$, denoted as $r_{il}(j,m)\in[-1,1]$ in \cite{hummel1983foundations}, is updated to $R_{il}(j,m)=0.5(r_{il}(j,m)+1)$. The original PRL-based updating becomes
\begin{equation}
    p_i^{(k+1)}(j)=\frac{p_i^{(k)}(j)\sum_{l=1}^{n_1}\sum_{m=1}^{n_2}R_{il}(j,m)p_l^{(k)}(m)}{\sum_mp_i^{(k)}(m)\sum_{l=1}^{n_1}\sum_{m=1}^{n_2}R_{il}(j,m)p_l^{(k)}(m)}
    \label{PRL_update}
\end{equation}
where $p_i(j)$ represents the probability that $i^\text{th}$ node of $\mathcal{V}_1$ matches $j^\text{th}$ node of $\mathcal{V}_2$. As discussed in the main paper, $p_i^{(k)}(j)$ in the numerator plays the role of weight factor. Specifically, if we ignore the effect of $p_i^{(k)}(j)$, Eq.~\eqref{PRL_update} becomes the power-iteration-like linear updating scheme commonly used in SM \cite{leordeanu2005spectral} and TM \cite{duchenne2011tensor}. 

Assume the highest probability for $i_0\in\mathcal{V}_1$ is $p_{i_0}(j_0)$. Hummel and Zucker proved that $p_{i_0}^{(k)}(j_0)$ consistently satisfies $\sum_{l=1}^{n_1}\sum_{m=1}^{n_2}R_{i_0l}(j_0,m)p_l^{(k)}(m)\geq p_{i_0}^{(k)}(j_0)$ during the updating if $\mathbf{R}$, the matrix form of $R_{i_1i_2}(j_1,j_2)$ for all $(i_1,i_2)$ and $(j_1,j_2)$, is symmetric \cite{hummel1983foundations}. To further accelerate the convergence, the weighting factor is replaced from $p_i^{(k)}(j)$ to $\sum_{l=1}^{n_1}\sum_{m=1}^{n_2}R_{il}(j,m)p_l^{(k)}(m)$, which leads to
\begin{equation}
    p_i^{(k+1)}(j)=\frac{[\sum_l\sum_mR_{il}(j,m)p_l^{(k)}(m)]^2}{\sum_j[\sum_l\sum_mR_{il}(j,m)p_l^{(k)}(m)]^2}
    \label{PRL_2}
\end{equation}
Define the compatibility coefficients as:
\begin{equation}
     R_{i_1i_2}(j_1,j_2)=
    \begin{cases}
        M_{i_1,j_1} & \text{$i_1=i_2$ and $j_1=j_2$} \\
        \hat{H}_{i_1,j_1,i_2,j_2} & \text{otherwise}
    \end{cases}
\end{equation}
where $M_{i,j}$ is the first-order compatibility between $i^\text{th}$ node of $\mathcal{V}_1$ and $j^\text{th}$ node of $\mathcal{V}_2$. $\hat{H}_{i_1,j_1,i_2,j_2}$ denotes the compatibility between $(i_1,i_2)\in\mathcal{E}_1$ and $(j_1,j_2)\in\mathcal{E}_2$. The numerator of Eq.~\eqref{PRL_2} becomes:
\begin{equation}
     \hat{p}_i^{(k+1)}(j)= (M_{i, j} p_i^{(k)}(j)+\sum_{l}\sum_m\hat{H}_{i,l,j,m}p_l^{(k)}(m))^2
     \label{PRL_4}
\end{equation}
$\hat{p}_i^{(k+1)}(j)$ is updated with the combination of first and second-order compatibilities. The main paper shows that for a consistent updating scheme, $\sum_{l}\sum_m\hat{H}_{i,l,j,m}p_l^{(k)}(m))^2\geq 0.25M_{i_0,j_0}^{-1}$.
By replacing the probability set $\mathbf{p}=\{p_i(j)\}$ with the vector $\mathbf{x}$, the column-wise flattening of soft-constraint assignment matrix $\mathbf{X}$, Eq.~\eqref{PRL_4} can be updated as
\begin{equation}
    \hat{\mathbf{x}}^{(k+1)}=(\hat{\mathbf{m}}\odot\mathbf{x}^{(k)} + \mathbf{H}\mathbf{x}^{(k)})^2
    \label{PRL_5}
\end{equation}
where $\odot$ is the element-wise multiplication and the first-order compatibility vector $\hat{\mathbf{m}}$ is obtained by column-wise flattening $\hat{\mathbf{M}}$. The square calculation in Eq.~\eqref{PRL_5} is also element-wise. $\mathbf{H}$ is the second-order compatibility matrix. Since both $\mathbf{H}$ and $\hat{\mathbf{m}}$ are dense, with the all-ones vector as $\mathbf{x}^{(0)}$, Eq.~\eqref{PRL_5} converges consistently.\par
In our work, the updating scheme for PRL-based hypergraph matching is extended to
\begin{equation}
    \hat{\mathbf{x}}^{(k+1)}= (\alpha\hat{\mathbf{m}}\odot\mathbf{x}^{(k)} + (1-\alpha)\mathcal{H}\otimes_1\mathbf{x}^{(k)}\otimes_2\mathbf{x}^{(k)})^2
    \label{PRLH}
\end{equation}
where $\otimes_l$ is the mode-$l$ product of the tensor and vector and $\alpha\in [0,1]$ is a balance weight between the first and third-order compatibilities. Since the third-order compatibility tensor $\mathcal{H}$ is highly sparse, the high value of $\hat{\mathbf{x}}^{(k+1)}$ in Eq.~\eqref{PRLH} is concentrated if the sparse tensor is reliable. In our work, a reliable tensor means most ground truth hyperedge pair compatibilities are successfully selected in the tensor blocks.

\section{Detailed Analysis in \cref{experiment}}

Due to the space limitation of the main paper, we provide a more detailed experiment analysis based on the results of \cref{experiment} to discuss the superiority and bottleneck of CURSOR.

\subsection{Memory Footprint Analysis in \cref{synthetic}}

\begin{table}
    \centering
    \caption{Detailed memory footprint of CURSOR in Tab.~\ref{tab-data} of the main paper.}
    \begin{tabular}{r|rrr|r}
    \hline
    Problem&\multicolumn{3}{c|}{Parameter}&Memory Footprint\\
    \hline
      $n_1$ vs $n_2$ & $t$ & $c$ & $r$ &$\mathbf{H}/(\mathbf{H}+\mathcal{H})$\\
      \hline
        30 vs 30 & 900 & 15 & 5 & 0.11MB/0.62MB\\
        30 vs 50 & 1500 & 15 & 5 & 0.20MB/0.82MB \\
        50 vs 50 & 2500 & 20 & 7 & 0.40MB/2.37MB\\
        50 vs 100 & 5000 & 100 & 10 & 3.97MB/9.62MB\\
        100 vs 100 & 10000 & 100 & 20 & 8.02MB/30.65MB\\
        300 vs 300 & 30000 & 200 & 20 & 0.14GB/0.21GB\\
        500 vs 500 & 50000 & 300 & 30 & 0.59GB/0.76GB\\
        800 vs 800 & 80000 & 400 & 50 & 1.95GB/2.02GB\\
        1000 vs 1000 & 100000 & 500 & 80 & 4.88GB/5.03GB\\
        \hline
    \end{tabular}
    
    \label{tab:memory}
\end{table}

Table~\ref{tab:memory} shows the detailed memory footprint of the experiment result with CURSOR in \cref{synthetic} of the main paper. Theoretically, the CUR decomposition of the matrix $\mathbf{H}$, requires $O(cn_1n_2)$ space complexity. The tensor $\mathcal{H}$, on the other hand, only needs $O(tr)$. For small-scale problems, the sparse tensor occupies most memory footprint with a small-size matrix. As the graph scale grows, with more columns selected from the compatibility matrices for higher matching accuracy, the main space occupation comes from $\mathbf{H}$, and the second-order CUR-based matching becomes the bottleneck for the graph matching problem. Although CURSOR can deal with larger-scale tasks compared to ANN, its capability to solve scalable problems is limited to the second-order matrix.

\begin{figure*}[htbp]
\centering

\subfloat[Source]{\includegraphics[width=0.25\textwidth]{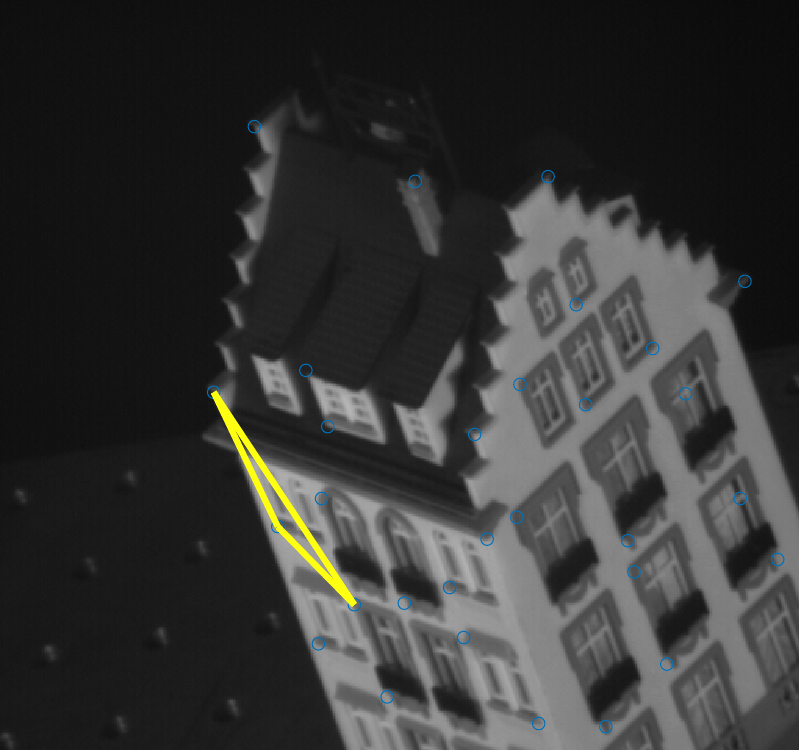}\label{subfig:source}}
\subfloat[ANN ($r_1=900$)]{\includegraphics[width=0.25\textwidth]{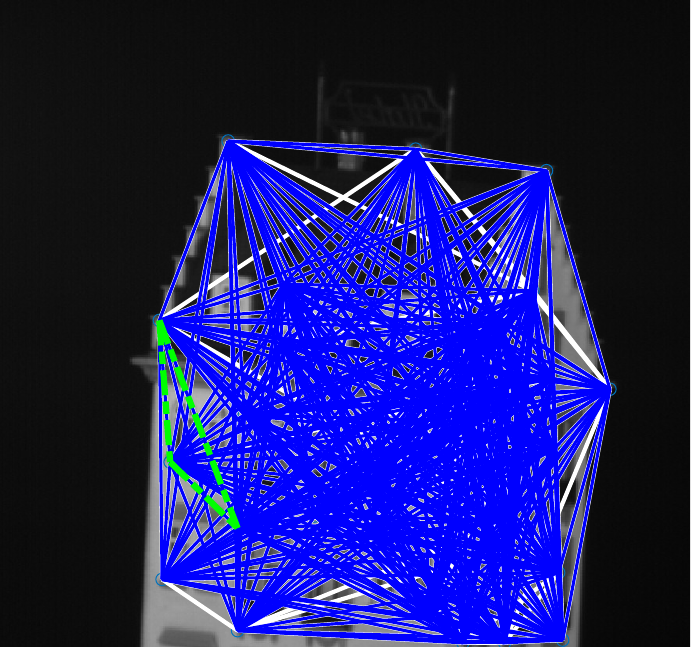}\label{subfig:ann}}
\subfloat[ANN ($r_1=50$)]{\includegraphics[width=0.25\textwidth]{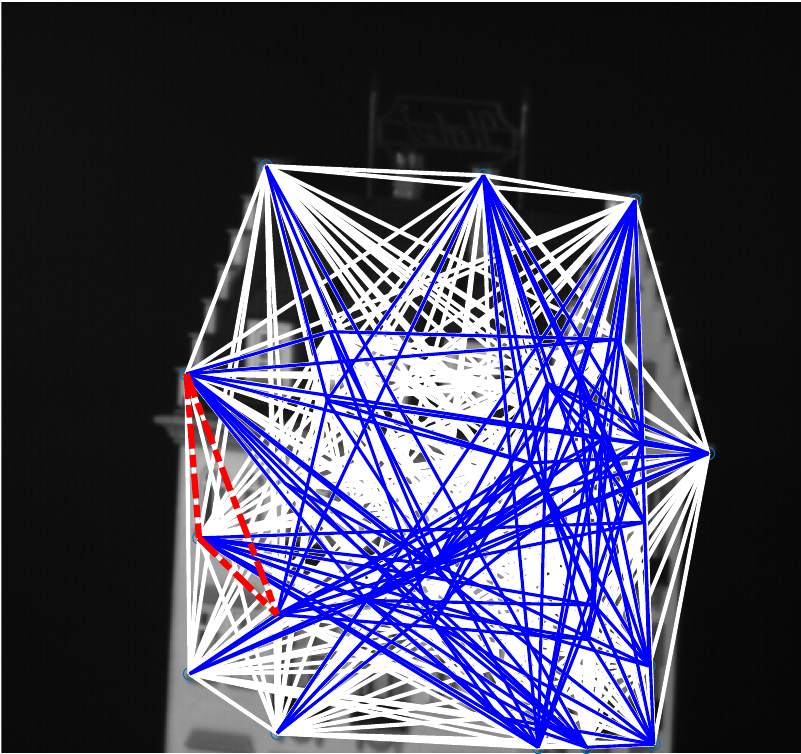}\label{subfig:ann50}}
\subfloat[CURSOR ($r=25$)]{\includegraphics[width=0.25\textwidth]{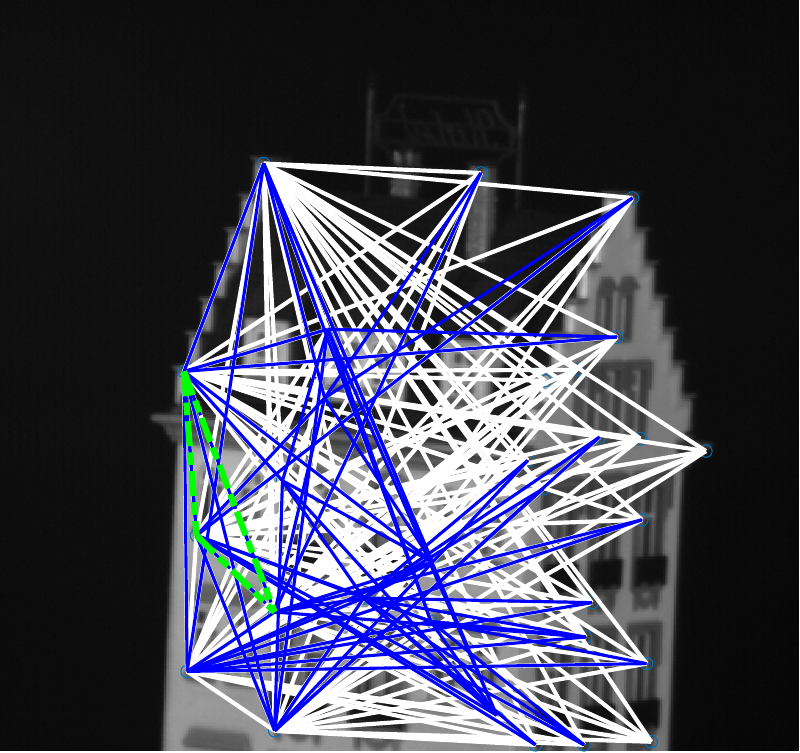}\label{subfig:cursor}}

\caption{The detailed hyperedge correspondences between one sampled hyperedge from the source (yellow triangle in \subref{subfig:source}) and target images with ANN (\subref{subfig:ann} and \subref{subfig:ann50}) and CURSOR \subref{subfig:cursor}. The white triangles in the target images denote all the hyperedges compared with the source hyperedge, and the blue triangles are the hyperedges with the highest compatibilities ($r_1$ for ANN and $r$ for CURSOR). The green dashed triangles represent the matched hyperedges, and the red one represents the mismatch.}
\label{fig:visual}
\end{figure*}

\subsection{Visualization analysis in \cref{house}} 
We provide a more detailed analysis of the experiment results from \cref{house} to demonstrate the superiority of CURSOR, as is shown in Fig.~\ref{fig:visual}. Traditional ANN-based methods compute the compatibilities between the sampled source hyperedges (yellow triangle in Fig.~\ref{subfig:source}) and all the target ones (fully connected white triangles in Figs.~\ref{subfig:ann}-\ref{subfig:ann50}). With the intermediate second-order result, CURSOR computes fewer compatibilities, as is shown in Fig.~\ref{subfig:cursor}. To find the corresponding hyperedge (green dashed triangles in Fig.~\ref{fig:visual}), a large amount (the hyperparameter $r_1=900$ in Fig.~\ref{subfig:ann}) of the highest compatibilities (blue triangles in Figs.~\ref{subfig:ann}-\ref{subfig:cursor}) should be selected for ANN. If we decrease $r_1$ to 50, some correct ones will be missed (red triangle in Fig.~\ref{subfig:ann50}). CURSOR can effectively find the hyperedges with the 25 highest compatibilities (green triangle in Fig.~\ref{subfig:cursor}). Compared to the traditional tensor generation methods, CURSOR can effectively increase the matching performance with less computational complexity.

\section{Ablation Studies}

To further analyze the effectiveness of several design choices in CURSOR, ablation studies were conducted on the 100-vs-110 random synthetic dataset introduced in \cref{synthetic} of the main paper. 

\subsection{CUR-based Pairwise Matching}

We first studied the effectiveness of the CUR-based second-order graph matching. Two experiments were designed to analyze the performance of the pairwise matching with the CUR decomposition of the second-order compatibility matrix $\mathbf{H}$ and the rough intermediate matching result, respectively. The dataset's noise level $\sigma$ varied in the range $[0,0.2]$.\par  

\begin{figure}
    \centering
    \subfloat[]{\includegraphics[width=0.24\textwidth]{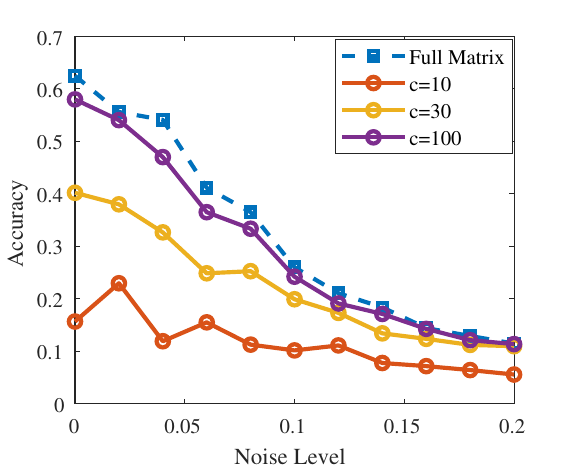}\label{fig:cur2}}
    \subfloat[]{\includegraphics[width=0.24\textwidth]{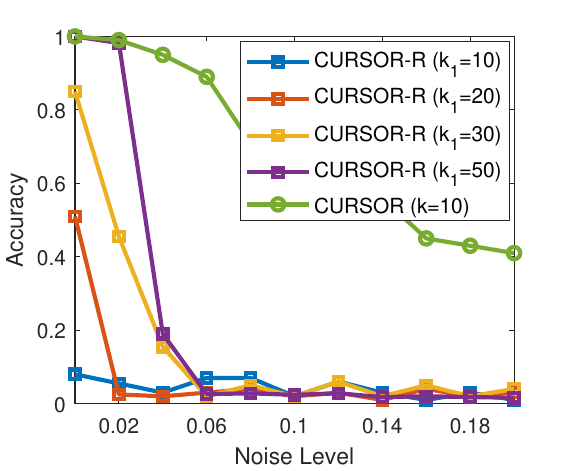}\label{fig:randomly}}
    \caption{\subref{fig:cur2} CUR-based second-order graph matching accuracy with various $c$. \subref{fig:randomly} The average accuracy using CURSOR with CUR-based pairwise matching result and CURSOR with randomly sampled indices (CURSOR-R).}
    \label{fig:pairwise_ab}
\end{figure}

\begin{table}
    \centering
    \caption{Average time consumption for second-order graph matching (in seconds)}
    \begin{tabular}{r|rr}
    \hline
    \multirow{2}{*}{Method}&\multicolumn{2}{c}{Time (s)}\\
    \cline{2-3}
         & Computing $\mathbf{H}$ & Matching\\
        \hline
        Full Matrix & 6.311 & 0.491\\
        
         $c=10$& 0.018 & 0.016\\
         $c=30$ & 0.031 & 0.016\\
        $c=100$ & 0.109 & 0.043\\
        \hline
    \end{tabular}
    
    \label{tab:second-time}
\end{table}

The CUR-based second-order graph matching was first evaluated on the dataset with various $c$, i.e., the number of randomly selected columns from $\mathbf{H}$. During the experiment, only the second-order graph-matching result was analyzed. We compared the proposed method with a PRL-based pairwise matching using the full compatibility matrix. The matching accuracy is shown in Fig.~\ref{fig:cur2}. Due to the low-rank estimation, the CUR-based method decreased the matching performance. Specifically, as the columns were randomly chosen, the result was quite poor when $c$ was relatively small. However, with less than $1\%$ (100 out of 100$\times$110) of columns selected, the CUR-based pairwise matching algorithm achieved a comparable result to the algorithm using the whole matrix. Table~\ref{tab:second-time} reports the average time consumption of the compatibility matrix generation and graph matching. With only a few columns calculated, the CUR-based second-order algorithm effectively accelerated the matching process.\par

The effectiveness of the intermediate second-order matching result was further studied. As discussed in the main paper, the second-order matching result $\mathcal{P}^k=\{\mathcal{P}_1^k,\cdots,\mathcal{P}_{n_1}^k\}$ consists $k$ best-matching target nodes for each source node, where $n_1$ denotes the number of the source nodes. The hypergraph matching result with $\mathcal{P}^k$ was compared to the result with $k_1$ randomly sampled indices in all three tensor modes (denoted as CURSOR-R). The parameter $c$ and $k$ from $\mathcal{P}^k$ for CURSOR with the pairwise matching result was set as 100 and 10 respectively. For CURSOR-R, $k_1$ varied from 10 to 50. For both methods, the number of randomly selected hyperedges from the source hyperedges $t=3000$, the number of highest compatibilities in each tensor block $r=100$, and the balance factor of PRL-based algorithm $\alpha$ was set to $0.2$. As shown in Fig.~\ref{fig:randomly},  CURSOR-R had a lower matching accuracy even for $k_1=50$. Assuming the number of nodes is $n_2$ in the target graph, theoretically, only 3 out of $3n_2^2$ fibers for each tensor block, i.e., key fibers in our work, contain the entry of the ground truth hyperedge pair. Due to the random sampling, the probability of selecting the key fiber is $(k_1/n_2)^2$, around $20.7\%$ when $k_1=50$ and $n_2=110$. Therefore, the sparse compatibility tensor generated by CURSOR-R was highly unreliable. CURSOR with the CUR-based pairwise matching result selected the key fiber in each tensor block with a high probability, effectively increasing the final accuracy.

\subsection{CURSOR vs ANN-based Tensor Generation}\label{sec:ANN}

\begin{figure*}[htbp]
\centering
\subfloat[]{\includegraphics[width=0.33\textwidth]{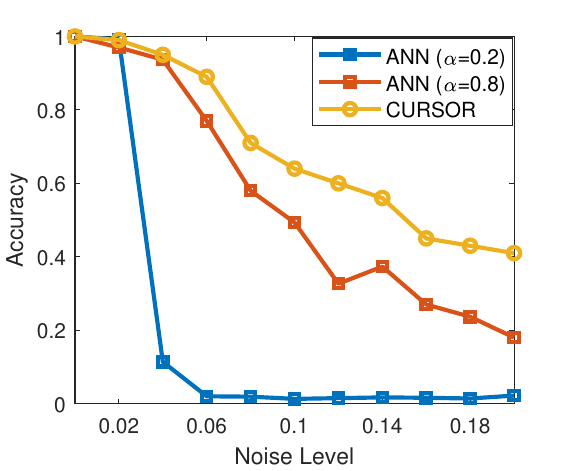}\label{fig:ann_acc}}
 \subfloat[]{\includegraphics[width=0.33\textwidth]{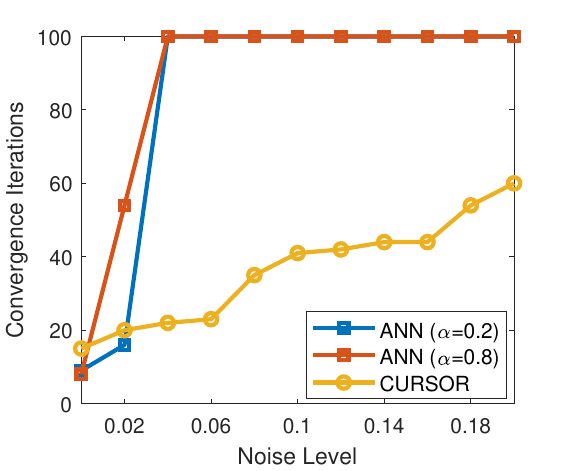}\label{fig:ann_i}}
 \subfloat[]{\includegraphics[width=0.33\textwidth]{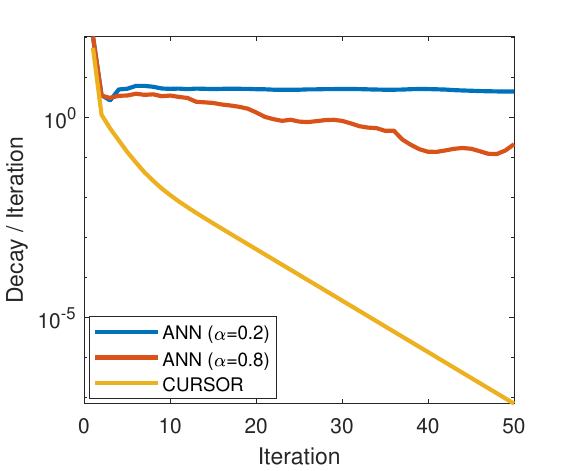}\label{fig:ann_c}}
\caption{Results on 100-vs-110 synthetic dataset comparing CURSOR with ANN, $\alpha=0.2$ for CURSOR. \subref{fig:ann_acc} The average matching accuracy using CURSOR and ANN with different $\alpha$. \subref{fig:ann_i} The average iterations for convergence using CURSOR and ANN with different $\alpha$. \subref{fig:ann_c} The decay $\|\mathbf{x}^{(k+1)}-\mathbf{x}^{(k)}\|_2$ per iteration with different $\alpha$ using CURSOR and ANN. The noise level $\sigma=0.1$.}
\label{fig:ablation_ann}
\end{figure*}

Experiment results in \cref{experiment} of the main paper have already shown that the proposed PRL-based algorithm with CURSOR achieved higher matching accuracy than other algorithms in most cases. One may wonder how the ANN-based tensor generation performs with the same hypergraph matching algorithm. In this experiment, we further compare CURSOR with the ANN-based tensor generation method applying PRL-based algorithm on the 100-vs-110 random synthetic dataset. During experiment, $c=100$, $k=10$ and $r=100$ for CURSOR. For ANN-based tensor generation, $r_1$ was set as 100 for the same tensor density. For both methods, $t=3000$ and $\alpha=0.2$. The stopping criterion of the PRL-based algorithm was set as $\|\mathbf{x}^{(k+1)}-\mathbf{x}^{(k)}\|_2\leq 10^{-8}$ and the maximum number of iteration was 100.\par 

The average matching accuracy is reported in Fig.~\ref{fig:ann_acc}. The ANN method shows an unstable performance with high $\sigma$ since the compatibility tensor was too sparse. When $\alpha$ was low, since the PRL-based algorithm focused on the third-order compatibilities, the ANN method did not generate a reliable compatibility tensor. To make the algorithm focus more on first-order compatibilities, $\alpha$ was further increased to 0.8 for the ANN case, which significantly improved the accuracy. The average number of iterations to converge is shown in Fig.~\ref{fig:ann_i}. When $\sigma>0.02$, the ANN method did not converge within 100 iterations. With the same tensor sparsity, CURSOR successfully converged in all the cases. The decay of $\mathbf{x}^{(k)}$ per iteration for the ANN method and CURSOR with $\sigma=0.1$ was further analyzed, as shown in Fig.~\ref{fig:ann_c}. The ANN method did not converge due to few ground truth hyperedge compatibilities selected from the whole tensor. However, CURSOR chose the non-zero compatibilities from a smaller reliable searching region. Therefore, it is capable of converging fast with the same tensor sparsity.

\section{Parameter Sensitivity Analysis}

Experiments below are provided to investigate how the hyperparameters in CURSOR affect the final results. Since the hyperparameter $t$, the number of randomly selected hyperedges, was thoroughly studied in previous works \cite{duchenne2011tensor, le2017alternating, lee2011hyper}, we do not redundantly analyze it here. The method was evaluated on the random 100-vs-110 synthetic dataset introduced in \cref{synthetic} of the main paper as well.

\subsection{Parameter in Second-order Graph Matching}

\begin{figure}[htbp]
\centering
 \subfloat[Hit rate over increasing $k$]{\includegraphics[width=0.24\textwidth]{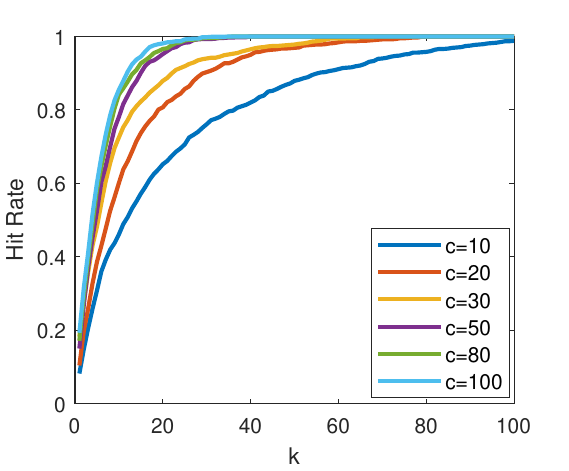}\label{fig:hit}}
 \subfloat[Accuracy with varying $c$]{\includegraphics[width=0.24\textwidth]{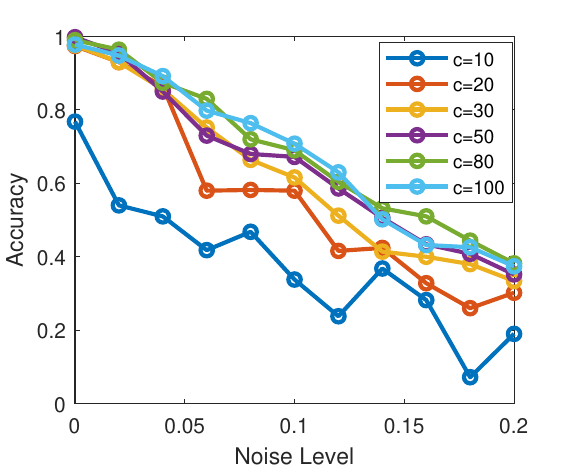}\label{fig:ab20}}
\caption{\subref{fig:hit} The average hit rate over increasing $k$ on a 100-vs-110 synthetic dataset with $\sigma=0.02$ from the CUR-based second-order graph matching results. \subref{fig:ab20} The average accuracy on the 100-vs-110 synthetic dataset with $\sigma\in[0,0.2]$ and $c\in[10,100]$.}
\label{fig:exp_second}
\end{figure}

The sensitivity of parameter $c$ to the whole framework was first analyzed. During the experiment, the whole compatibility matrix was calculated directly with noise level $\sigma=0.02$. CUR decomposition was evaluated on the matrix with various $c$ for the pairwise graph matching. To analyze the influence of $c$ on the second-order matching result, we define hit rate, calculated as $\sum_{i=1}^{n_1}\delta_i/{n_1}$, where
\begin{equation}
    \delta_i=
    \begin{cases}
        1 & \text{The true match $j\in\mathcal{P}_i^k$} \\
        0 & \text{Otherwise}
    \end{cases}
\end{equation}
The average hit rate was computed over the increasing number of selected highest compatibilities $k$, as shown in Fig.~\ref{fig:hit}. Each curve represents the hit rate with $c$ during CUR decomposition. To achieve the same hit rate and set $k$ as small as possible, theoretically, the number of selected columns needed to be as large as possible. However, when increasing $c$ from 50 to 100, the gain on hit rate is minor, with a relatively small $k$ to reach a promising hit rate like 0.9. %However, with a linearly increased computation cost, $k$ decreased little if $c$ was increased from 50 to 100 for a promising hit rate, like 0.9. 
Therefore, to balance the time consumption and the matching performance, a relatively small proportion of columns is sufficient for the rough pairwise matching result. \par 

The influence of $c$ on the final matching result was further analyzed. During the experiment, we set $k=10$, $r=100$, and $\alpha=0.2$. The noise level of the dataset was assigned as $\sigma\in[0,0.2]$. The result is reported in Fig.~\ref{fig:ab20}. Since the columns of the compatibility matrices were randomly selected, the matching result was unstable when $c$ was less than 20. The performance gradually saturated with over 50 columns (around $0.5\%$ of the total columns). The experiment results in Fig.~\ref{fig:hit} show that 50-100 columns can already provide a reliable second-order matching result for the following higher-order process, effectively decreasing the computation cost.

%We further analyze the influence of the second-order graph matching result by implementing it on the House dataset with 20-vs-30 point problems. We set $t=3n_1n_2$ and $r=n_1n_2$ for the ANN-based tensor generation method. Then we directly select the corresponding fibers with the result of the CUR-based second-order matching algorithm on the tensor blocks to simulate the compatibility tensor generated by the CUR-based method. $c$ is set as 10. The matching result with RRWHM and ADGM is shown in Fig.~\ref{fig:exp_second}~\subref{fig:ab20}. The matching result using the non-zero elements of the corresponding fibers leverages the one using all the non-zero elements generated by the ANN-based method. Therefore, it can be proved that the fibers selected by the CUR-based second-order graph-matching results are useful. We can select the nearest neighbors in these fibers instead of the whole tensor block.

\subsection{Parameters During Tensor Generation}

\begin{figure}[htbp]
\centering
\subfloat[Accuracy with varying $k$]{\includegraphics[width=0.24\textwidth]{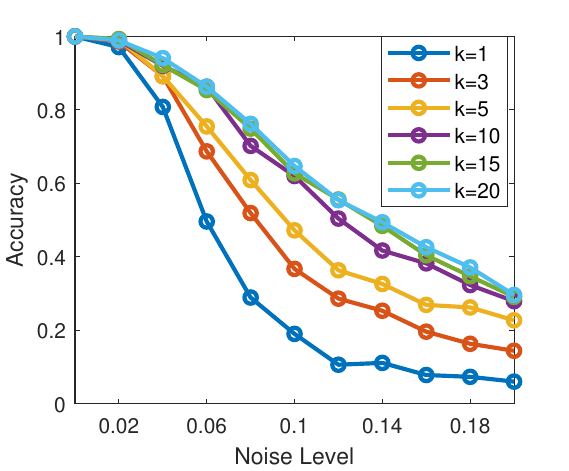}\label{fig:k_acc}}
 \subfloat[Accuracy with varying $r$]{\includegraphics[width=0.24\textwidth]{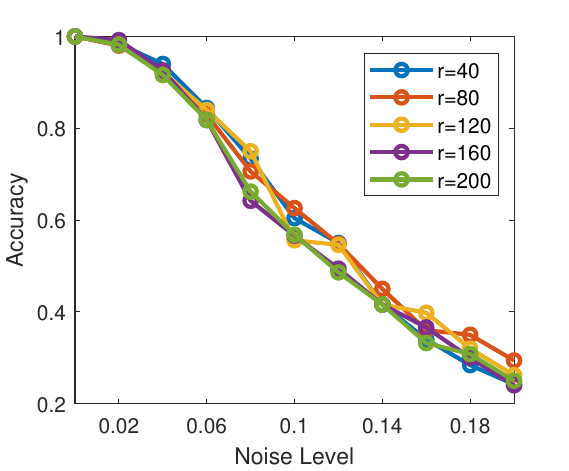}\label{fig:r_acc}}
\caption{Results on 100-vs-110 synthetic dataset with $\sigma\in[0,0.2]$ using CURSOR. \subref{fig:k_acc} The matching accuracy with increasing $k$. \subref{fig:r_acc} The matching accuracy with increasing $r$.}
\label{fig:exp_third}
\end{figure}

\begin{figure*}[htbp]
\centering
\subfloat[]{\includegraphics[width=0.33\textwidth]{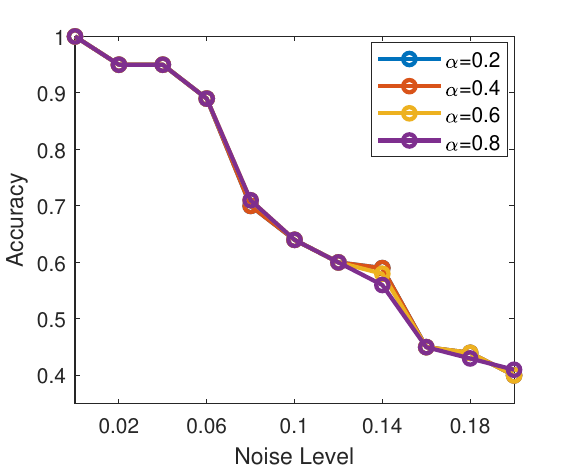}\label{fig:prl_a}}
 \subfloat[]{\includegraphics[width=0.33\textwidth]{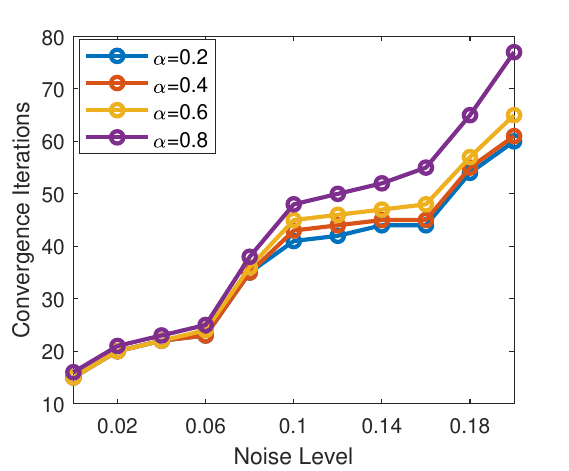}\label{fig:prl_i}}
 \subfloat[]{\includegraphics[width=0.33\textwidth]{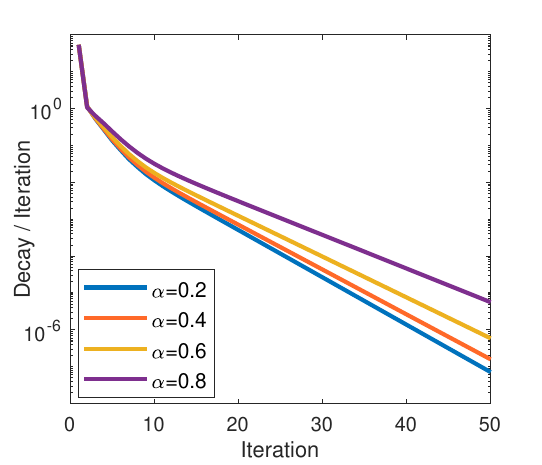}\label{fig:prl_c}}
\caption{Results on 100-vs-110 synthetic dataset with PRL-based matching algorithm. \subref{fig:prl_a} The average matching accuracy using CURSOR with different $\alpha$. \subref{fig:prl_i} The average iterations for convergence using CURSOR with different $\alpha$. \subref{fig:prl_c} The decay $\|\mathbf{x}^{(k+1)}-\mathbf{x}^{(k)}\|_2$ per iteration with different $\alpha$ settings using CURSOR. The noise level $\sigma=0.1$.}
\label{fig:ablation_prl}
\end{figure*}

The sensitivity of $k$ and $r$ to the final matching accuracy was further studied. During the experiment, $c$ was set as 100 and $\alpha=0.2$.\par 
To figure out the influence of $k$, the parameter $r$ was set as 100, and $k$ varied from 1 to 20. The result is shown in Fig.~\ref{fig:k_acc}. It is obvious that a higher $k$ can achieve higher robustness for target points with high noise impact. However, with a polynomial $O(tk^2n_2)$ computation cost for tensor generation, the matching performance gradually reached its peak. For instance, the matching accuracy increased more than $35\%$ from $k=1$ to $k=5$ when $\sigma=0.08$, but less than $2\%$ from $k=15$ to $k=20$. For each tensor block, if the entry of the ground truth paired hyperedge compatibility was included in $r$ non-zero elements, the corresponding nodes would be matched with a high probability. Therefore, an appropriate $k$ needs to be set to balance the computation cost and the performance in practical use.\par

The effect of $r$ was further analyzed by setting $k$ as 10 and varying $r$ from 20 to 200, as shown in Fig.~\ref{fig:r_acc}. Unlike the results in Fig.~\ref{fig:k_acc}, CURSOR with the highest $r$ performed the worst. The reason may be that when the ground truth hyperedge pair is already included in the non-zero compatibilities, more redundant compatibilities can cause lower matching performance.\par

\subsection{Parameters of PRL-based Matching Algorithm}

To study the sensitivity of $\alpha$, the matching accuracy and convergence speed using CURSOR were further analyzed. During the experiment, $c=100$, $k=10$, and $r$ was set as $100$. The stopping criteria of the PRL-based algorithm was assigned as $\|\mathbf{x}^{(k+1)}-\mathbf{x}^{(k)}\|_2\leq 10^{-8}$. The matching accuracy with various $\alpha$ is shown in Fig.~\ref{fig:prl_a}. With a reliable compatibility tensor, the PRL-based matching algorithm achieved almost the same performance regardless of $\alpha$. We further analyzed their convergence speeds, as shown in Figs.~\ref{fig:prl_i} and \ref{fig:prl_c}. As the noise level increased, more iterations were required to satisfy the stopping criteria, and a lower $\alpha$ achieved faster convergence. As discussed in the previous sections, the parameter $\alpha$ is a balanced factor between first and third-order compatibilities. Although the method converged faster with $\alpha=0$, the first-order compatibilities stabilized the matching process and increased the matching performance under some extreme circumstances.\par

%\section{Memory Footprint Analysis}

%Finally, the memory consumption of the hypergraph matching with CURSOR is analyzed. Theoretically, the CUR decomposition of the matrix $\mathbf{H}$, requires $O(cn_1n_2)$ space complexity. The tensor $\mathcal{H}$, on the other hand, only needs $O(tr)$. Table~\ref{tab:memory} shows the detailed memory footprint of the experiment result with CURSOR in \cref{synthetic} of the main paper. As the graph scale grows, the main space occupation comes from $\mathbf{H}$, and the second-order CUR-based matching becomes the bottleneck for a larger-scale graph matching problem.

%{
%    \small
%    \bibliographystyle{ieeenat_fullname}
%    \bibliography{main}
%}

\end{document}